\documentclass[preprint]{elsarticle}

\usepackage{lineno,hyperref,amssymb,amsmath, amsthm,graphicx,tabularx,subfigure,multirow,caption}

\usepackage[dvipsnames]{xcolor}

\newtheorem{theorem}{Theorem}[section]

\newtheorem{lemma}[theorem]{Lemma}
\journal{Journal of Pattern Recognition}
\begin{document}

\begin{sloppypar}
\begin{frontmatter}

%
% Title
%
\title{
Elucidating Robust Learning with Uncertainty-Aware Corruption Pattern Estimation
}
% 
%
% Authors
%
\author[alpha]{Jeongeun Park }
\ead{baro0906@korea.ac.kr}
\author[alphb]{Seungyoun Shin}
% \ead{2018112005@dongguk.edu}
\author[alphc]{Sangheum Hwang}
\author[alpha]{Sungjoon Choi\corref{cor1}}%\fnref{fn1}}
\ead{sungjoon-choi@korea.ac.kr}
\cortext[cor1]{Corresponding author}
% \cortext[cor2]{Principal corresponding author}
% \fntext[fn1]{This is the specimen author footnote.}
% \fntext[fn2]{Another author footnote, but a little more longer.}
\address[alpha]{Department of Artificial Intelligence, Korea University, Seoul 02841, Korea}
\address[alphb]{Department Computer Engineering, Dongguk University, Seoul 04620, Korea}
\address[alphc]{Department of Data Science, Seoul National University of Science and Technology, \\
Seoul 01811, Korea}

%
% Abstract
%
\begin{abstract}
Robust learning methods aim to learn a clean target distribution from noisy and corrupted training data where a specific corruption pattern is often assumed a priori. Our proposed method can not only successfully learn the clean target distribution from a dirty dataset but also can estimate the underlying noise pattern. To this end, we leverage a mixture-of-experts model that can distinguish two different types of predictive uncertainty, aleatoric and epistemic uncertainty. We show that the ability to estimate the uncertainty plays a significant role in elucidating the corruption patterns as these two objectives are tightly intertwined. We also present a novel validation scheme for evaluating the performance of the corruption pattern estimation. Our proposed method is extensively assessed in terms of both robustness and corruption pattern estimation in the computer vision domain. Code has been made publicly available at \url{https://github.com/jeongeun980906/Uncertainty-Aware-Robust-Learning}.
\end{abstract}

%
% Keywords
%
\end{frontmatter}

%
% Introduction
%
\section{Introduction}

In this paper, we focus on the problem of robust learning \cite{choi_20,han_18,patrini_17,yu_19,wei_20,ekambaram_16,bootkrajang_14,lu_15} with its emphasis on elucidating the corruption patterns on the noisy training dataset. Most existing robust learning studies \cite{han_18,wei_20} assume that the label corruption pattern is solely a function of class information, also known as Class-Conditional Noise (CCN). While this CCN assumption is simple to formulate, it may not be useful in practice in that it is more natural to assume that the noise pattern is a function of input instances which is often referred to as an Instance-Dependent Noise (IDN) learning problem \cite{berthon_20}. 

However, the original IDN learning problem is likely to be infeasible in that it has to estimate a $C \times C$ class transition matrix per input instance. Due to this intractability, recent work on the IDN setting focuses on a simple binary classification problem \cite{cheng_20_3,bootkrajang_20} or requires a small clean dataset \cite{berthon_20}. To mitigate this issue, we cast the IDN problem into a two-stage problem of first partitioning the input space using uncertainty measures and then estimating the label transition matrix per each group, which will be referred to as a Set-Dependent Noise (SDN) learning problem. In particular, a specific type of predictive uncertainty, named aleatoric uncertainty, is used to partition the input space. The clusters with high aleatoric uncertainty can be viewed as collective outliers \cite{Karamcheti_21}, a subset of inputs with a specific noise pattern. 

We would like to stress that robust learning and uncertainty estimation problems are intimately related to each other as robust learning deals with the noisy training data, which naturally gives rise to predictive uncertainty. The predictive uncertainty can be decomposed into epistemic and aleatoric uncertainty. The former focuses on the reducible part of the uncertainty (model uncertainty), which may come from the lack of training data. In contrast, the latter comes from the irreducible part (data uncertainty), such as the measurement noise. Our proposed method can estimate both types of uncertainty in a unified framework, and it plays a significant role in achieving robustness and estimating the SDN patterns. 

To this end, we utilize a mixture-of-experts model for classification tasks named mixture logit networks (MLN) and present an effective training method to achieve both robustness and explainability by revealing the label corruption process. Although a mixture-of-experts method was first presented in the 80s, we show its effectiveness on a robust learning framework with simple modifications and show that it can also estimate the noise distributions as well. We first present an uncertainty estimation method for the MLN that can distinguish two different types of predictive uncertainties, epistemic (model) uncertainty and aleatoric (data) uncertainty. Then, the estimated uncertainty is utilized for the uncertainty-aware regularization method. Intuitively speaking, unlike a single deterministic model (e.g., a ResNet), using the MLN allows us to model multi-modal (and possibly noisy) target distributions, which plays a crucial role in achieving both robustness as well as explainability. Furthermore, we present an evaluation scheme on SDN settings, which gives information about the collective outliers and label noise distribution of sets. 

The main contributions of this work are threefold. 1) We propose a simple yet effective robust learning method leveraging a mixture-of-experts model on various noise settings. 2) The proposed method can not only robustly learn from noisy data but can also discover the underlying set-dependent noise pattern (i.e., the noise transition matrix) as well as the two types of predictive uncertainties (i.e., aleatoric and epistemic uncertainty) within the dataset. 3) Finally, we present a novel evaluation scheme for validating the set-dependent corruption pattern estimation performance.

%
% Related Work 
% 
\section{Related Work} \label{sec:rel}

In the context of robust learning, the label noise patterns can be roughly categorized into two groups, Class-Conditional Noise (CCN) and Instance-Dependent noise (IDN) settings, based on which information the label corruptions are made. Note that the IDN setting is much more practical as it is more natural to assume that the label noise pattern is a function of inputs. Furthermore, the IDN setting can inherently incorporate the CCN setting. Our proposed method can cope with both CCN and IDN settings. While most of the robust learning literature focuses on simply estimating the clean target distribution, a number of works have been recently made on achieving both robustness and the ability to estimate the label noise patterns.

One possible approach is to extract a clean subset, then utilize the set to learn the clean target signal. Co-Teaching \cite{han_18} utilizes two separate networks (a teacher network and a student network) by teaching the student network using the teacher network. Co-Teaching+ \cite{yu_19} extends Co-Teaching by further leveraging the disagreement strategy. JoCoR~\cite{wei_20} is based on Co-Teaching+, which uses a joint loss function of minimizing cross-entropy while maximizing the agreement between two networks to achieve better robustness. DivideMix \cite{li_20} leverages a hybrid approach by fitting the loss using a GMM to divide the dataset into the clean labeled set and noisy unlabeled set, then utilizes MixMatch\cite{berthelot_19} with divided sets for training two separate networks. WarPI \cite{sun_22} achieves its robustness by adaptively rectifying the training procedure for the classification network within the meta-learning scenario.
%Small loss tricks of providing small risk for unseen clean data have been used in \cite{wang_19,ma_20,lyu_19}. 

The other approach is learning the label transition matrix (noise patterns) from noisy data to estimate the clean-class posterior. F-correction \cite{patrini_17} estimates the noise transition matrix and applies it to loss function correction. Dual-T \cite{yao_20} incorporates a matrix factorization method to avoid directly estimating a noisy class posterior without any anchor points (i.e., clean data). Total variation regularization \cite{zhang_21_3} effectively regularizes the predicted probability to be more distinguishable by restricting the total variance distance resulting in a better estimation of the noise transition matrix. 

The IDN setting is more practical than the CCN setting in that it is more natural to assume that the annotator gets confused by ambiguous instances leading to mislabeling. Only recently, a few papers have incorporated confidence estimation and noise transition matrix prediction to handle confusing instances. However, most of the work only uses the confidence estimation to robustly learn clean target distribution in instance-dependent noise settings. Cheng et al. \cite{cheng_20} present confidence regularization to prevent overfitting in multi-class classification problems. \cite{cheng_20} is further extended in \cite{cheng_20_2} by designing a sample sieve method to get clean instances from the noisy dataset, using confidence regularized cross-entropy loss. The confidence regularized method gives information about the corruption of each instance but does not provide information about label noise patterns. Another approach is to estimate the noise transition matrix instance-wise to correct the loss function. Part-dependent noise (i.e., PDN) was introduced by Xia et al. \cite{xia_20}  which approximates the transition matrix by the combination of transition matrices for each instance. Yang et al.\cite{yang_21} first collect a predicted clean set to learn the noise transition matrix and then train a classifier with a corrected loss function based on the estimated noise transition matrix.

% Instance-dependent
Perhaps, the most similar setting to ours is \cite{berthon_20} which introduced a confidence-scored Instance-Dependent Noise (IDN) setting; a label noise is given based on prior information about confidence score by annotators. It uses both confident estimation and noise transition matrix estimation for robust learning. Then, the model utilizes the corrected loss function using the confidence-based noise transition matrix. However, the proposed Set-Dependent Noise (SDN) setting differs from the confidence-scored Instance-Dependent Noise (IDN) setting in that the annotators can divide the set by the ambiguity measure of each instance, making it more intuitive and straightforward. Furthermore, the proposed method directly estimates both corruption information and the noise distribution without the necessity of training multiple networks. The categorization of robust learning papers is shown in Table \ref{tab:paper}

% Table: Related Work
%
\begin{table}[t]
{ 
%\small
\centering
\resizebox{0.9\columnwidth}{!}{%
\begin{tabular}{|c|c|c|}
    \hline
    & Class-Conditional Noise (CCN) 
    & Instance-Dependent Noise (IDN) 
    \\
    \hline
    Robsut Learning 
    & \cite{han_18,yu_19,wei_20,wang_19,ma_20,lyu_19,choi_20,li_20} & \cite{berthon_20,cheng_20} 
    \\
    \hline
    + Noise Transition Matrix Prediction 
    & \cite{xia_19,yao_20} 
    & \cite{xia_20,wang_21}, Ours
    \\
 \hline
\end {tabular}%
}
\caption{
Categorization of robust learning papers
}
}
\end {table} \label{tab:paper} 

% Mixture Models
The mixture-of-experts models have been widely used in robust learning \cite{shao_19,irie_18,choi_20}. SsSMM \cite{shao_19} incorporates a student-teacher model similar to MentorNet \cite{han_18}, but employs a finite mixture models for student networks, updating via an EM algorithm \cite{mclachlan_07} in semi-supervised manner. For robust learning for language domains, Irie et al. \cite{irie_18} proposed a recurrent adaptive mixture model to represent diverse outputs. ChoiceNet \cite{choi_20} utilizes a mixture density network to model the correlated outputs where the correlation between the target and noisy distributions is estimated in an end-to-end manner. Our proposed method is also based on a mixture-of-experts model; however, a novel uncertainty-aware regularization method is presented. 

%
% Problem Formulation
%
\section{Problem Formulation} \label{sec:prob}

%
% Training Data Generation Process
%
\subsection{Training Data Generation Process}

% Main setting
In this paper, we focus on the classification task of finding a mapping from an instance $\mathbf{x}$ (e.g., an RGB image) to an output $\mathbf{y}$ (e.g., an one-hot vector) where the input $\mathbf{x}$ and the output $\mathbf{y}$ are sampled from the input distribution $p(\mathbf{x})$ and the clean target distribution $p(\mathbf{y}|\mathbf{x})$. We assume that some noise patterns can be induced to both input and output where we denote $\tilde{\mathbf{x}}$ and $\tilde{\mathbf{y}}$ as the corrupted input and output, respectively. The input noise pattern, $\tilde{\mathbf{x}} \sim p(\tilde{\mathbf{x}},\mathbf{x})$,can be adding more blur to the instance so that the resulting image is obscured or applying artificial manipulation to the image (e.g., CutMix \cite{yun_19}). 

% Output corruptions
Roughly speaking, the output corruption process can be divided into twofold: the Class-Conditional Noise (CCN) and the Instance-Dependent Noise (IDN) settings. For the CCN setting, it is assumed that the training label information is corrupted via a single label transition matrix $T \in \mathbb{R}^{C \times C}$ where $C$ is the number of classes and $[T]_{ij} = p(\mathbf{y}_j | \mathbf{y}_i)$ is the probability of a label $i$ being shifted to a label $j$. For example, we can simply select a certain portion of the training data and shuffle the labels uniformly randomly or shift the labels by assigning label $1$ to $2$, label $2$ to $3$, and so forth. In the robust learning literature, the formal and the latter are often referred to as symmetric and asymmetric noise patterns, respectively. On the other hand, the ICN assumes that the noise pattern is a function of an instance (i.e., $T(\mathbf{x}) \in \mathbb{R}^{C \times C}$). However, as a single instance can only have a single target label, it would be unrealistic to have the whole label transition matrix $T$ per instance. 

% Set-dependent
Throughout this paper, a Set-Dependent Noise (SDN) setting is utilized where we assume that the training dataset is partitioned into subsets where each subset contains its own label corruption matrices. This assumption is rather more practical in that it is more natural to assume that the annotators will be more likely to make mistakes on a specific subset consisting of hard instances. We would like to stress that our proposed method can estimate the label noise patterns in both CCN and SDN settings without the necessity of additional clean data.

%
% Robust Learning and Corruption Pattern Estimation
%
\subsection{Robust Learning and Corruption Pattern Estimation}

The main objectives of the proposed method are twofold: the first is to robustly learn the underlying clean target signal out of noisy training data, and the other is to gain the explainability of the prediction via estimating the label corruption information as well as the predictive uncertainty. Specifically, we disentangle the total uncertainty into aleatoric and epistemic uncertainty similar to Kendall et al. \cite{kendall_17} and will be explained in the next section. Aleatoric uncertainty corresponds to the \textit{irreducible} part of the uncertainty, which is inherent in the data generation process (e.g., measurement noise). On the other hand, epistemic uncertainty captures the model uncertainty, which may \textit{reduce} as we have more training data. 

With respect to the label corruption information, we estimate the Set-Dependent Noise (SDN) pattern of the training dataset without the necessity of a clean validation dataset. Note that the SDN inherently handles the CCN as it can simply condition the whole data. Specifically, we estimate the label transition matrix conditioned on the subset of training of test data where the corruption rates and the noise patterns, symmetric or asymmetric, can be estimated from the transition matrix. 

%
% Proposed Method
%
\section{Proposed Method} \label{sec:proposed}

%
% Figure: General process of the proposed method
%
\begin{figure}[!t]
    \includegraphics[width=1\textwidth]{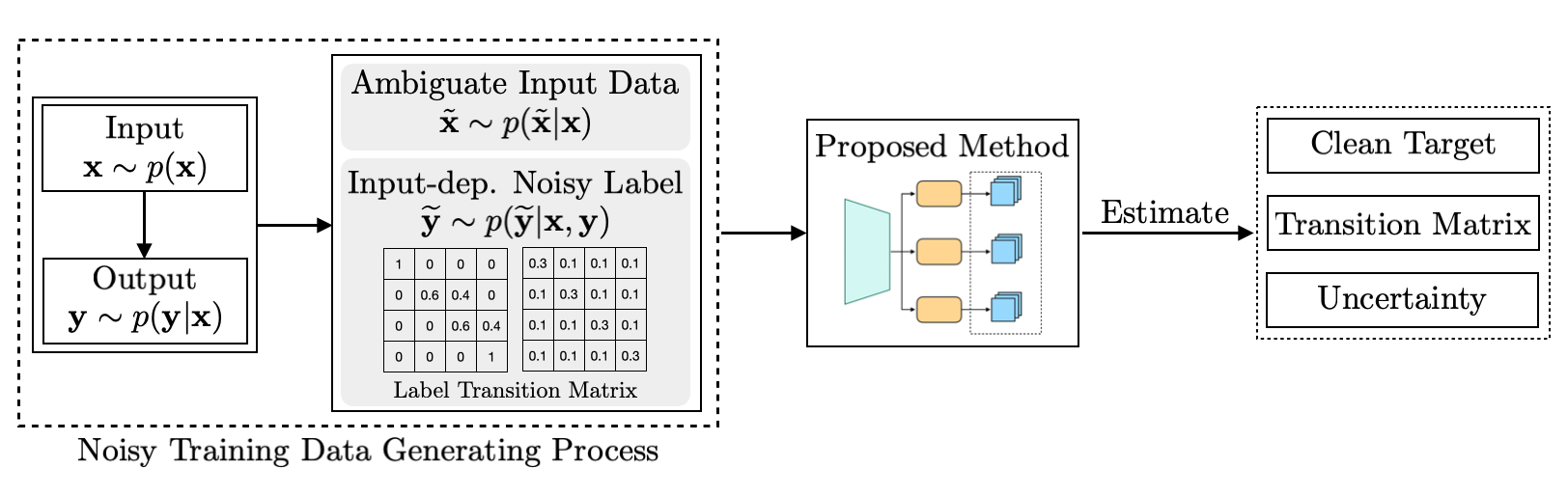}
    \caption{General process of the proposed method}
    \label{fig:objective}
\end{figure}

% What do we propose 
We present a robust learning method via a mixture-of-experts model for a classification task named mixture logit networks (MLN) and a noise pattern estimation method utilizing the outputs of the MLN. To fully utilize the multiple mixtures, we further propose an uncertainty-aware regularization method. We empirically show that this regularization method plays an influential role in achieving both robustness and explainability. The intuition behind leveraging the mixture model is that, when given corrupted training data, the noise pattern will give rise to the discrepancy of the prediction outputs, where a single deterministic model (e.g., a ResNet) often fails to correctly capture the clean target signal. However, as a mixture model, when adequately trained, can better capture the inconsistent output patterns (including both clean and noisy distributions), it not only can robustly learn the underlying target distribution but also can model the noise patterns injected in the data generating process. The overall process of the proposed method is illustrated in Figure \ref{fig:objective}.

\begin{figure}[!t]
    \centering
    \includegraphics[width=0.9\textwidth]{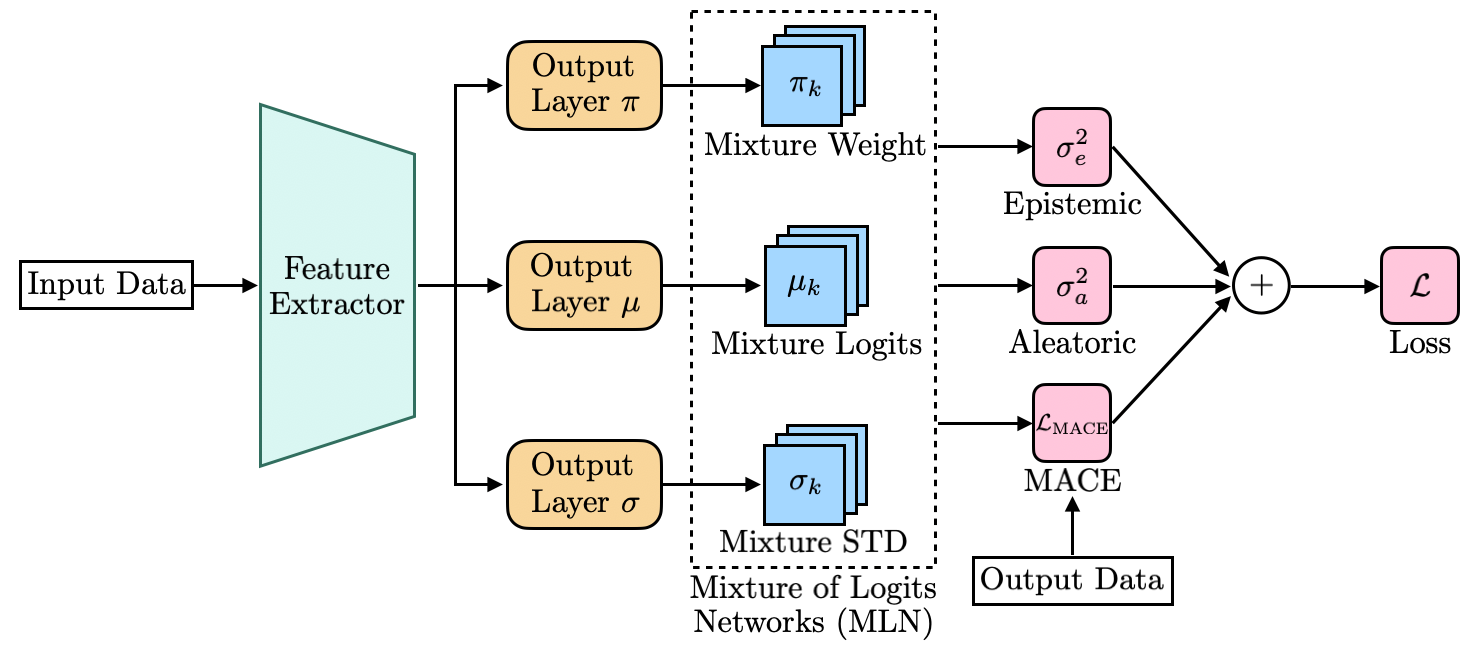}
    \caption{The proposed mixutre of logit network architecture}
    \label{fig:arc}
\end{figure}

% Architecture
The MLN architecture is illustrated in Figure \ref{fig:arc}. Suppose that the number of mixtures is $K$, then the MLN outputs consist of mixture weights $\{\pi_k \}_{k=1}^{K}$, logits $ \{\boldsymbol{\mu}_k \}_{k=1}^K$ where $\boldsymbol{\mu}_k \in \mathbb{R}^C$ and $C$ is the number of classes, and Mixture standard deviations (Mixture STD) $\{\sigma_k \}_{k=1}^K$. Note that only the uppermost layer is modified. Hence the total number of parameters does not change significantly. 

%
% Figure: Mixture Logit Architecture
%

%
% Subsection: uncertainty estimation 
%
\subsection{Uncertainty Estimation using the MLN}

We first present ways to estimate two types of uncertainties with the MLN: epistemic (model) uncertainty and aleatoric (data) uncertainty. We denote $\sigma_e$ as epistemic uncertainty and $\sigma_a$ as aleatoric uncertainty. Fist, epistemic uncertainty is computed as follows.
\begin{equation}
    \sigma_e^2 =\sum_{j=1}^K \left(\sum_{c=1}^C  \pi_j \left\| \mu^{(c)}_j(\mathbf{x})-\sum_{k=1}^K \pi_k \mu^{(c)}_k(\mathbf{x})\right\|^2 \right)
    \label{eq:epis}
\end{equation}
where $\mu^{c}_k$ is logit of label $c$ in $k$th mixture.

On the other hand, epistemic uncertainty ($\sigma_e$) indicates how much the model is uncertain about its prediction. (\ref{eq:epis}) corresponds to the weighted average variance of each mixture's predicted logits, which can be seen as disagreements between $\{\boldsymbol{\mu}_k \}_{k=1}^K$.On the other hand, aleatoric uncertainty ($\sigma_a$) is computed as follows.
\begin{equation}
    \sigma_a^2 =\sum_{k=1}^K \pi_k \sigma_k (\mathbf{x})\label{eq:alea}
\end{equation}
Aleatoric uncertainty captures noise inherent in observation and how much the model is uncertain about its data. (\ref{eq:alea}) indicates the weighted average of each mixture's predicted STD of the given input. Mixture STD $\{ \sigma_k \}_{k=1}^K$ denotes predicted noise by mixtures, also can be used as attenuation factor for loss function, similar to Kendal et al. \cite{kendall_17}.

%
% Mixture of the Attenuated Losses
%
\subsection{Mixture of the Attenuated Losses}

We present a Mixture of the Attenuated Cross-Entropy (MACE) loss for effectively training the MLN. We denote the target as $y_i$, which can either be clean or noisy (i.e., $\tilde{y_i}$) depending on the dataset. The proposed loss function consists of cross-entropy loss divided by the standard deviation of each mixture (i.e., loss attenuation) and then a weighted summation of the attenuated loss for each mixture. Mixture standard deviation $\{ \sigma_k \}_{k=1}^K$ corresponds to the expected measurement noise of each instance.

The MACE loss function is defined as follows:
\begin{equation}
   \mathcal{L}_{MACE} = \frac{1}{N}\sum_{i=1}^N\sum_{k=1}^K \pi_k(\mathbf{x}_i)\frac{l({\boldsymbol{\mu}_k(\mathbf{x}_i)},y_i)}{\sigma_k(\mathbf{x}_i)} \label{eq:MACE}
\end{equation}
where $l(\boldsymbol{\mu}_k(\mathbf{x}_i),y_i)$ is the cross entropy loss.

If an input is ambiguous or corrupted, it becomes more likely to make a false prediction. Then $\sigma_k$ will increase to reduce the overall loss function of the prediction. As a result, this attenuated factor prevents the overfitting of the model to the corrupted dataset, making the proposed model more robust.

%
% Uncertainty-aware Regularization Method
%
\subsection{Uncertainty-aware Regularization Method}

%
% Figure: Effect of the Regularizer On half-moon dataset
%
\begin{figure}[!h]
% \subfloat[Symmetric Noise]{%
% \begin{minipage}[t]{1\linewidth}
    \centering
    \includegraphics[width=0.33\textwidth]{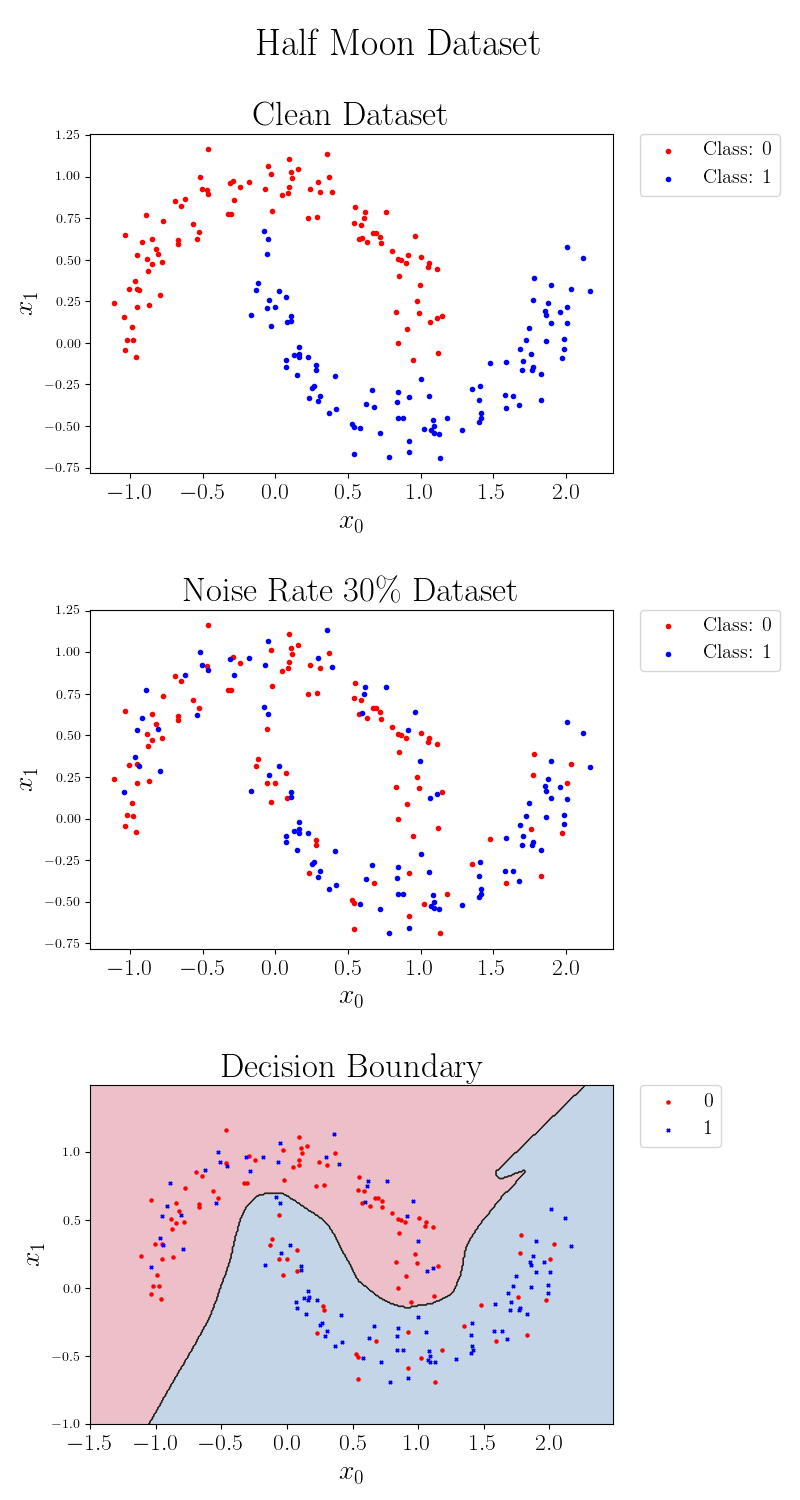}\hfill
    \includegraphics[width=0.33\textwidth]{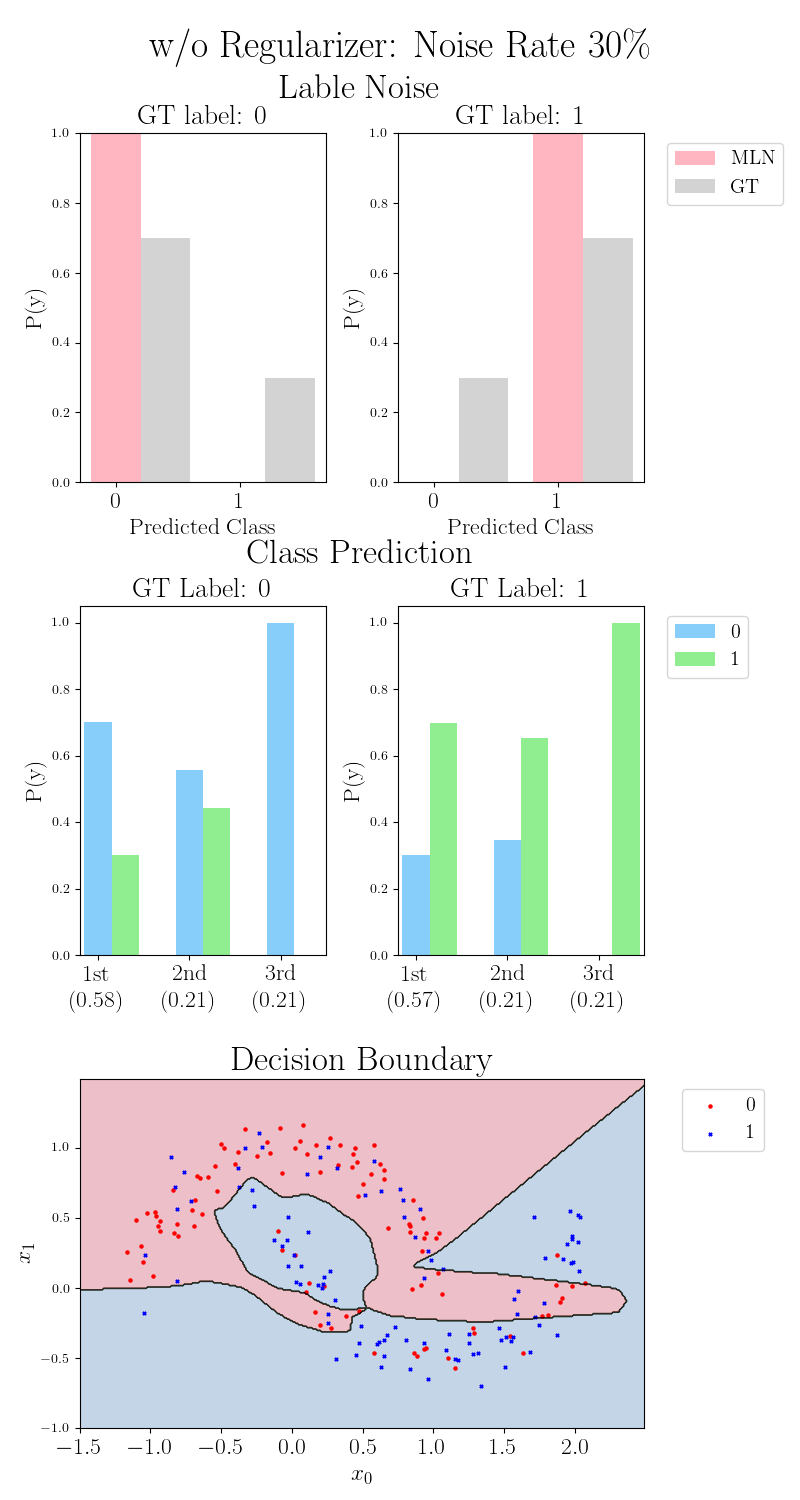}\hfill
    \includegraphics[width=0.33\textwidth]{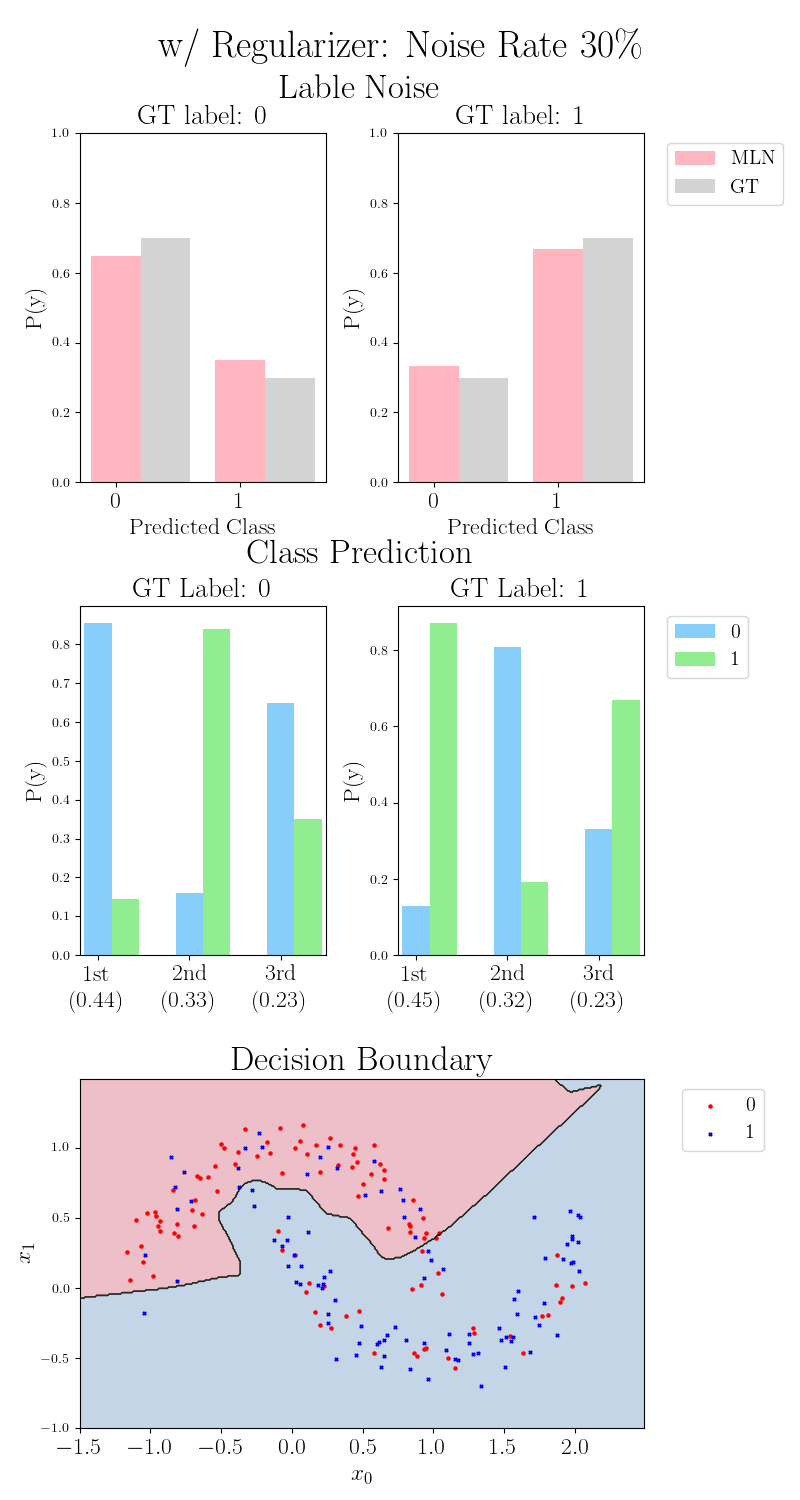}%
% \end{minipage}%
%   }\par
\caption{Effect of the Regularizer On half-moon dataset}
\label{fig:toy_reg}
\end{figure}

We observe that there could exist two possible problems when training only with the original MACE loss function: a simple increment of the mixture standard deviation will minimize the loss and the insufficient usage of mixtures. To resolve these issues, we propose a novel regularization method utilizing predictive uncertainty measures. Let us first present the proposed loss function utilizing the uncertainty measures:
\begin{equation}
    \mathcal{L}(\mathcal{D})= \mathcal{L}_{MACE} -\lambda_1 \sigma_e + \lambda_2 \sigma_a. \label{eq}
\end{equation}

The first problem is that $\{ \sigma_k \}_{k=1}^K$  in (\ref{eq:MACE}) will simply grow to minimize the total loss. To prevent this, we need the regularization term, $\{ \sigma_k \}_{k=1}^K$. This is inspired by Kendall et al. \cite{kendall_17}, where the loss function is based on the Gaussian likelihood and is the sum of attenuated regression loss and regularization of $\{ \sigma_k \}_{k=1}^K$. We present a parameter $\lambda_2$ as a weight.  

Next, it is known that the mixture-of-experts model is prone to use only one or few components, incapable of capturing the various data distributions. Moreover, we observe that different mixtures are easily agreeable with each other, which is inappropriate to represent a multi-modal distribution. Due to this property, the whole model often fails to learn both clean and corrupted data distribution, leaving the model vulnerable to noisy data. To this end, we regularize epistemic uncertainty to be large, which naturally incentives to utilize more mixtures where $\lambda_1$ is the weight parameter.

We illustrate effect of this regularizer in Figure \ref{fig:toy_reg}. The synthetic dataset consists of two-dimensional inputs (i.e., $x_0$ and $x_1$) for a binary classification problem where we assign label $0$ to instances on the upper moon and label $1$, otherwise. We corrupt the label information by flipping the labels at the rate of 30\% and training the MLN with three mixtures. In the first column, we illustrate the clean half-moon dataset, noisy half-moon dataset, and decision boundary trained by the clean half-moon dataset. The second and third columns present the effect of the regularizer. The first row presents the estimated flipping rate, computed from (\ref{eq:temperature}), which will be discussed in the later section. The result indicates that the proposed regularizer helps the better prediction of the noise distribution. The second row of Figure \ref{fig:toy_reg} shows that the output of each mixture disagrees with each other, making a better representation. Furthermore, the third row illustrates that the proposed regularizer smooths the decision boundary in the presence of outliers. 

%
% Corruption Pattern Estimation
%
\subsection{Corruption Pattern Estimation}

% Noise Transition Matrix
We further use the output of the MLN to gain information about the noise corruption pattern. This can be done by estimating the noise transition matrix. The noise transition matrix $T_{ij}(\mathbf{x})$ \cite{xia_19,yao_20} indicates the probabilities of the clean labels flipping to noisy labels. The notation represents the probability that the instance $\mathbf{x}$ with the clean label $y=i$ will have a noisy label $\tilde{y}=j$. Formally, the noise transition matrix is defined as follows.
\begin{equation}
    T_{ij}(\mathbf{x}) = P(\tilde{y}=j|y=i,\mathbf{x}) \label{eq:def_tm}
\end{equation}

% The distribution of label noise for each instance as a multi-modal distribution. By this assumption, noise distribution can be understood as a weighted sum between a clean distribution and corrupted distribution. 

Capturing multi-modality is one of the strengths that the mixture model possesses. This property leads the MLN to model a multi-modal distribution on a noisy instance, representing both the clean and noisy label distribution. As the label corruption patterns can be regarded as a multi-modal distribution, we introduce an auxiliary random variable $z$ to estimate the noise transition matrix. 

\begin{equation}
    T_{ij}(\mathcal{D}_l;z) =\frac{1}{|\mathcal{D}_l(i)|}\sum_{\mathbf{x} \in \mathcal{D}_l(i)} P(\tilde{y}=j|\mathbf{x})
\end{equation}
here, $\mathcal{D}_l$ denotes the set indexed $l$ and $\mathcal{D}_l(i) = \{\mathbf{x}|y=i,(\mathbf{x},y) \in \mathcal{D}_l\}$. 
For SDN setting set index can be 0 (clean) or 1 (ambiguous) and for CCN set index will be 0 (total). 
\begin{equation}
    T_{ij}(\mathcal{D}_l;z) =\frac{1}{|\mathcal{D}_l(i)|}\sum_{\mathbf{x} \in \mathcal{D}_l(i)}\sum_{k=1}^{K}{P(z=k)P(\tilde{y}=j|z=k,\mathbf{x}})\label{eq1}
\end{equation}
where K is total number of mixtures, z is latent variable, and $P(z=k)$ is weight of component distribution denoted as $\pi_k$ above.

Starting from (\ref{eq1}), we denote the $\widehat{P}(\mathbf{y}|\mathbf{x})$ as the the soft-max output vector approximating $P(\mathbf{y}|\mathbf{x})$ by parametrized model and estimator for the noise transition matrix as $\widehat{T}$. In addition, since we cannot observe clean $y$, we assume $y$ as Bayes optimal labels, the class labels that maximize the clean class posteriors $ \hat{f}(\mathbf{x}) := \underset{y}{\text{argmax}} \hat{P}(y|\mathbf{x})$. Furthermore, we define a subset $\widehat{\mathcal{D}}_l(i)$ with the Bayes optimal label index $i$ in the set $\widehat{\mathcal{D}}_l$, as we do not have any prior knowledge about the clean $y$. 
\begin{equation}
    \widehat{\mathcal{D}}_l(i) = \{\mathbf{x}|\hat{f}(\mathbf{x})=i,\mathbf{x} \in \mathcal{D}_l\}\label{eq:set}
\end{equation}
Then, the (\ref{eq1}) can be modified as follows.
\begin{equation}
    \widehat{T}_{ij}(\mathcal{D}_l;z) =\frac{1}{|\widehat{\mathcal{D}}_l(i)|}\sum_{\mathbf{x} \in \widehat{\mathcal{D}}_l(i)} \sum_{k=1}^{K}{\widehat{P}(z=k)\widehat{P}(\tilde{y}=j|z=k,\mathbf{x})} \label{eq:lemma}
\end{equation}
The Lemma \ref{lemma:tm} shows (\ref{eq:lemma}) is a valid transition matrix, since the row-wise sum of predicted transition matrix becomes one.
%
% Lemma
%
\begin{lemma} \label{lemma:tm}
The row wise sum of the proposed transition matrix estimation in (\ref{eq:lemma}) becomes one.
\begin{equation}
    \sum_{j=1}^C{\widehat{T}_{ij}(D_l;z)}=1
\end{equation}
\begin{proof} 
The row wise sum of the proposed estimated noise transition matrix becomes as follows.
\begin{equation}
    \sum_{j=1}^C\widehat{T}_{ij}(\mathcal{D}_l;z) =\sum_{j=1}^C\frac{1}{|\widehat{\mathcal{D}}_l(i)|}\sum_{\mathbf{x} \in \widehat{\mathcal{D}}_l(i)} \sum_{k=1}^{K}{\widehat{P}(z=k)\widehat{P}(\tilde{y}=j|z=k,\mathbf{x})}
\end{equation}
Since $j$ is not dependent to $|\widehat{\mathcal{D}}_l(i)|$ and  $\widehat{P}(z=k)$, the equation can be rewritten as follows.
\begin{equation}
    \sum_{j=1}^C\widehat{T}_{ij}(\mathcal{D}_l;z) =\frac{1}{|\widehat{\mathcal{D}}_l(i)|}\sum_{\mathbf{x} \in \widehat{\mathcal{D}}_l(i)} \sum_{k=1}^{K}{\widehat{P}(z=k)\sum_{j=1}^C\widehat{P}(\tilde{y}=j|z=k,\mathbf{x})}
\end{equation}
By the definition of the categorial distribution and mixture weight, $\sum_{j=1}^C\widehat{P}(\tilde{y}=j|z=k,\mathbf{x})=1$ and $\sum_{k=1}^{K}\widehat{P}(z=k)=1$. 
\begin{equation}
    \sum_{j=1}^C\widehat{T}_{ij}(\mathcal{D}_l;z) =\frac{1}{|\widehat{\mathcal{D}}_l(i)|}\sum_{\mathbf{x} \in \widehat{\mathcal{D}}_l(i)}1 = 1
\end{equation}
As we define the set $\widehat{\mathcal{D}}_l(i)$ as (\ref{eq:set}), the proposed noise transition matrix holds $\sum_{j=1}^C{\widehat{T}_{ij}(D_l;z)}=1$.
\end{proof}
\end{lemma}

As the confidence of the softmax-output decreases on the noisy dataset, the predicted transition matrix often suffers from being too smooth. To better estimate the noise transition matrix, inspired by Liang et al. \cite{liang_17}, we apply temperature scaling for the softmax activation. We set the temperature to zero, which makes the softmax function an indicator function, driving to the predicted noise transition matrix inherent to the confidence score. $\hat{P}_{\text{scaled}}$ is defined as follows.

\begin{equation}
    \hat{P}_{\text{scaled}}(\tilde{y}=j|z=k,\mathbf{x}) = \mathbb{I}_j(\underset{c}{\text{argmax}}\hat{P}(\tilde{y}=c|z=k,\mathbf{x})) \label{eq:temperature}
\end{equation}

The scaled transition matrix is defined by replacing $\hat{P}$ to $\hat{P}_{\text{scaled}}$ in (\ref{eq:lemma}).

% \revise{After obtaining noise transition matrix, we can get noise rate as well.}

% \revise{Given noise transition matrix $\widehat{T}_{ij}$, noise rate is defined as follows.
% \begin{equation}
%     \text{noise-rate} \ (\widehat{T}_{ij}) = 1 - \frac{1}{C}\sum^C \widehat{T}_{ii}
%     % \frac{1}{|\mathcal{D}|}\sum_{\mathbf{x} \in \mathcal{D}} P(\tilde{y}|\mathbf{x}) = 
% \end{equation}

% We define noise rate as one minus average of diagonal element of noise transition matrix.
% }
%
% Experiments
%
\section{Experiments}

In this section, we present experimental results of validating the robustness of the proposed method. We first describe the implementation details for experiment settings, including datasets, corruption patterns, and hyperparameters. Next, we present the results in the CCN setting and compare them with benchmarks. Furthermore, we utilize the estimated uncertainty measures to distinguish the collective outliers in the SDN settings, where the noise transition matrices of each partition are estimated and compared with the ground truth. 

%
% Implementation Details
%
\subsection{Implementation Details}
%
% CCN Datasets
%
\paragraph{Class-Conditional Noise (CCN) Setting}
We first construct a Class-Conditional Noise (CCN) dataset with clean instances and noisy labels whose corruption rate is solely a function of class information. We evaluate the proposed method on four different datasets, MNIST, CIFAR10, and CIFAR100. These datasets are popularly used for evaluating the robustness of the image classification algorithms. Following JoCoR~\cite{wei_20}, we conduct experiments on four different label corruption patterns: Symmetry-20\%, Symmetry-50\%, Symmetry-80\%, and Asymmetry-40\%. 

%
% SDN Datasets
%
\paragraph{Set-Dependent Noise (SDN) setting}

A Set-Dependent Noise (SDN) setting is utilized where the dataset is partitioned into two subsets: clean set and ambiguous set. In particular, we define an ambiguous set as a set containing a pair of corrupted instances and noisy labels. We experiment on two different datasets: Dirty-MNIST and Dirty-CIFAR10. Dirty-MNIST, proposed by Mukhoti et.al.~\cite{mukhoti_21} is formed as the union of MNIST set and Ambiguous-MNIST set. Ambiguous MNIST contains corrupted instances where it has multiple plausible labels but contains only one GT label. To conduct set-dependent noise on the Dirty-MNIST dataset, we added label noise on the Ambiguous-MNIST set. We validate on four different label noise patterns: Symmetry-20\%, Symmetry-50\%, Symmetry-80\%, and Asymmetry-40\%. We define the Dirty-CIFAR10 dataset, which contains half of the original CIFAR10 dataset and the other half ambiguated with the CutMix~\cite{yun_19} method. We choose the CutMix method to maintain the scheme that samples on ambiguous sets should have multiple possible labels but has one GT label. Again, to form the SDN setting, we added label noise on the ambiguous set.

%
% Hyperparameters
%
\paragraph{Hyperparameters}

We use a three-layer CNN for MNIST and a seven-layer CNN for both CIFAR10 and CIFAR100 following JoCoR~\cite{wei_20}. For the Clothing1M dataset, we use ResNet50 as a backbone. We set the batch size as 128 and use an Adam optimizer with the learning rate $10^{-3}$ and train the model with 200 epochs for CIFAR10, CIFAR100, and 20 epochs for MNIST. The learning rate is decayed 0.2 times for every ten epochs for CIFAR10, and CIFAR100 and 0.2 decay rate for every five epochs on the MNIST dataset. On Clothing1M, we use a SGD optimizer with a learning rate of $10^{-3}$, decaying 0.1 times for every 30 epochs. The weight decay rate is set to $10^{-3}$, and we train the model for 80 epochs. We set the minimum of $\{\sigma_k\}_{k=1}^K$ as one and maximum as ten by using a sigmoid function except for the IDN setting and MLN+MixUp on Section \ref{sec:ssl}. In these settings, where $\{\sigma_k\}_{k=1}^K$ is set the minimum as 0.1 and maximum as 1. In addition, we set the number of the mixtures to be $20$ for all experiments, which should be large enough to cover all the noise distribution. In addition, when using MixUp \cite{zhang_17} augmentation, we tune $\alpha$ as four. Furthermore, we set regularizer hyperparameters as $\lambda_1=1$, $\lambda_2=1$ except for CIFAR100, where we scale the parameter to $\lambda_1=0.1$, $\lambda_2=1$, and the IDN setting where $\lambda_1=10$.
Regularizer parameters are selected using cross-validation results. On cross-validation, we assume there exists a small clean set and use 10\% of the clean test set as the validation set.

%
% Robust Learning Performance
%
\subsection{Class-Conditional Noise (CCN) Setting}
This section first shows the robust learning performance of MNIST, CIFAR10, and CIFAR100 in CCN settings and compares them with supervised-learning methods. Next, we prove the validity of the estimated noise transition matrix on the CCN setting. Furthermore, as state-of-the-art models deploy semi-supervised methods to deal with noisy labels, we combine the proposed method with a semi-supervised method to observe the effectiveness of the proposed method.

\subsubsection{Robust Learning Accuracy}
We conduct robust learning experiments with the Class-Conditional Noise (CCN) setting to investigate the performance of the MLN. We evaluate the test accuracy on four datasets with four different noise patterns and compare with Noise Adaptation \cite{goldberger_16}, F-correction \cite{17_patrini}, Co-teaching \cite{han_18}, Co-teaching+ \cite{yu_19} and JoCoR \cite{wei_20}. The test accuracy on MNIST is shown in Table \ref{tab:mnist_acc}. The proposed method outperforms on the Symmetry-80\% setting, and on other noise settings, it is compatible with the compared methods. However, the results on CIFAR10 in Table \ref{tab:cifar10_acc} show that our method outperforms the compared methods on Symmetry-80\% and Asymmetry-40\%, with the second-best performance on other noise patterns. Furthermore, Table \ref{tab:cifar100_acc} presents the test accuracy on CIFAR100 dataset. The MLN outperforms on Symmetry-80\% and Asymmetry-40\% noise patterns and performs second-best on the rest of the noise patterns. We would like to note that the proposed method shows its strengths in heavy corruptions, such as Symmetry-80\% and Asymmetry-40\%. The proposed method shows a significant performance margin on large corruption rates, such as Symmetry-80\% and Asymmetry-40\%. 
% \revise{Furthermore, we added MixUp \cite{zhang_17} regularization with the MLN and showed a performance increase in all noise settings except for CIFAR10 Asymmetry 40\% noise rate. We observe that the proposed method can be used with other regularization methods for robust learning like MixUp \cite{zhang_17} leading to a performance increase.}

%
% Table: 
%
\begin {table}[!t]
\caption {MNIST Test Accuracy in CCN setting} 
\label{tab:mnist_acc} 
\resizebox{\columnwidth}{!}{%
\begin{tabular}{c||c|c|c|c}
\hline
Method & Symmetry-20\% & Symmetry-50\% & Symmetry-80\% & Asymmetry-40\% \\
\hline \hline
Noise Adaptation~\cite{goldberger_16} & 89.0 $\pm$ 0.05 & 98.75 $\pm$ 0.05 & 14.08 $\pm$ 0.59 & 60.08 $\pm$ 0.01 \\
\hline
F-correction~\cite{patrini_17} & 98.37$\pm$0.28 & 95.70$\pm$0.6& 85.33$\pm$2.28 & 95.19$\pm$1.29\\
\hline
Co-teaching~\cite{han_18} & 99.08$\pm$0.04 &98.19$\pm$0.09 & 85.26$\pm$0.11 & 96.69$\pm$0.70\\
\hline
Co-teaching+~\cite{yu_19}& 99.00$\pm$0.10 & \textbf{98.83}$\pm$0.11 & 86.24$\pm$0.16 & \textbf{98.65}$\pm$0.18\\
\hline
JoCoR~\cite{wei_20} & \textbf{99.20}$\pm$0.07 & 98.76$\pm$0.06 &86.00$\pm$0.15 & 98.46$\pm$0.21 \\ 
\hline
MLN (ours) & 98.97$\pm$0.01 & 98.32$\pm$0.01 & \textbf{93.49} $\pm$ 0.21 & 97.13$\pm$0.10\\
\hline \hline
\end {tabular}%
}
% }
\end {table}

%
% Table: 
%
\begin {table}[!t]
\caption {CIFAR10 Test Accuracy in CCN setting} \label{tab:cifar10_acc} 
\resizebox{\columnwidth}{!}{%
\begin{tabular}{c||c|c|c|c}
\hline
Method & Symmetry-20\% & Symmetry-50\% & Symmetry-80\% & Asymmetry-40\% \\
\hline \hline
Noise Adaptation \cite{goldberger_16} & 82.03 $\pm$ 0.1 & 39.88 $\pm$ 0.21 & 10.0 $\pm$ 0.0 & 72.44 $\pm$ 0.1 \\
\hline
F-correction\cite{patrini_17} & 68.74$\pm$0.20 & 42.71$\pm$0.42 & 15.88$\pm$0.42 & 70.60$\pm$0.40\\
\hline
Co-teaching\cite{han_18} & 78.23$\pm$0.27 & 71.30$\pm$0.13 & 26.58$\pm$2.22 & 73.78$\pm$0.22\\
\hline
Co-teaching+\cite{yu_19}& 78.71$\pm$0.34 & 57.05$\pm$0.54 & 24.19$\pm$2.74 & 68.84$\pm$0.20\\
\hline
JoCoR\cite{wei_20} & \textbf{85.73}$\pm$0.19 & \textbf{79.41}$\pm$0.25 & 27.78$\pm$3.06 & 76.36$\pm$0.49\\ 
\hline

MLN (ours) & 84.20$\pm$0.05 & 77.88$\pm$0.07 & \textbf{41.83}$\pm$0.10 & \textbf{76.62}$\pm$0.07\\
% \hline

\hline \hline
\end {tabular}%
}
\end {table}

%
% Table
%
\begin {table}[!t]
\caption {CIFAR100 Test Accuracy in CCN setting} \label{tab:cifar100_acc} 
\resizebox{\columnwidth}{!}{%
\begin{tabular}{c||c|c|c|c}
\hline
Method & Symmetry-20\% & Symmetry-50\% & Symmetry-80\% & Asymmetry-40\% \\
\hline \hline
Noise Adaptation \cite{goldberger_16} & 30.14 $\pm$ 0.09 & 2.83 $\pm$ 0.05 & 1.0 $\pm$ 0.0 & 22.23 $\pm$ 0.1\\
\hline
F-correction\cite{patrini_17} & 37.95$\pm$0.10 &24.98$\pm$1.82 & 2.10$\pm$2.23 & 25.94$\pm$0.44\\
\hline
Co-teaching\cite{han_18} &43.73$\pm$0.16 & 34.96$\pm$0.50 & 15.15$\pm$0.46 & 28.35$\pm$0.25\\
\hline
Co-teaching+\cite{yu_19}& 49.27$\pm$0.03 & 40.04$\pm$0.70 & 13.44$\pm$0.37 & 33.62$\pm$0.39\\
\hline
JoCoR\cite{wei_20} & \textbf{53.01}$\pm$0.04 & \textbf{43.49}$\pm$0.46 & 15.49$\pm$0.98 & 32.79$\pm$0.35 \\ 
\hline
MLN (ours) & 51.60$\pm$0.08 & 42.22$\pm$0.07 &\textbf{19.88}$\pm$0.14\ & \textbf{36.36}$\pm$0.10\\

 \hline \hline
\end {tabular}%
}
\end {table}

%
% Table
%
% \begin {table}[!t]
% \caption {TREC Test Accuracy} \label{tab:trec_acc} 
% \centering
% \resizebox{0.5\columnwidth}{!}{%
% \begin{tabular}{c|c|c }
%  \hline \hline
%  Flipping Rate & Baseline\cite{kim_14} & MLN (ours) \\
%  \hline \hline
% Symmetry-20\% & 84.88$\pm$0.14 & \textbf{87.38}$\pm$0.12\\
% \hline
% Symmetry-50\% & 57.28$\pm$0.20 & \textbf{74.79}$\pm$0.90\\
% \hline
% Symmetry-70\% & 29.14$\pm$0.61 & \textbf{55.78}$\pm$1.39\\
% \hline
% Asymmetry-40\% & 72.63$\pm$0.96 & \textbf{74.17}$\pm$0.28\\
%  \hline \hline
% \end {tabular}%
% }
% \end {table}

%
% Class-Conditional Noise (CCN) Transition Matrix Estimation
%
\subsubsection{Noise Transition Matrix Estimation}

In this section, we evaluate the noise transition matrix estimation on the Class-Conditional Noise (CCN) setting. We estimate the noise transition matrix using (\ref{eq:temperature}), which is an anchor-free method that does not require a clean validation set. We evaluate the noise transition matrix on MNIST and CIFAR10 datasets with Symmetry-20\%, Symmetry-50\%, Symmetry-80\%, Asymmetry-40\%, Dual-40\%, and Tridiagonal-60\% noise patterns \footnote{We call Dual for two mislabeled classes and Tridiagonal for three mislabeled classes}.

%
% Table
%
\begin {table}[!t]
\caption {Evaluation of Noise Transition Matrix ATV in CCN setting (average total variation)(x100), KTD(Kendall-Tau distance)} \label{tab:tm} 
\resizebox{\columnwidth}{!}{%
\begin{tabular}{c c|c c|c c}
 \hline \hline
 & & \multicolumn{2}{c}{Noise Adaptation\cite{goldberger_16}} &
\multicolumn{2}{c}{MLN(Ours)} \\
\multirow{4}{*}{MNIST} &
 Noise Rate & ATV & KTD & ATV & KTD \\
 \cline{2-6}
 & Symmetry-20\% & 22.19 & 0.4472 & \textbf{15.40} & 0.4472\\
 \cline{2-6}
 & Symmetry-50\% & 48.67 & 0.4472 & \textbf{8.46} & 0.4472\\ 
 \cline{2-6}
 & Symmetry-80\% & 3.41 & 0.4472 & \textbf{2.33} & 0.4472\\ 
 \cline{2-6}
 & Asymmetry-40\% & 13.36 & 0.5237 & \textbf{10.19} & 0.5349\\
%  \cline{2-6}
%  & Dual-40\% & &  & 13.64 & \textbf{0.4940}\\
%  \cline{2-6}
%  & Tridiagonal-60\% & &  & 10.96 & \textbf{0.5937}\\
 \hline
 \multirow{4}{*}{CIFAR10} 
 & Symmetry-20\% &23.15 & 0.4472 & \textbf{20.71} & 0.4472\\
 \cline{2-6}
 & Symmetry-50\% & 34.36 & 0.4472 & \textbf{9.69} & 0.4472\\ 
 \cline{2-6}
 & Symmetry-80\% & 6.86 & 0.4472 & \textbf{5.85} & 0.4472\\ 
 \cline{2-6}
 & Asymmetry-40\% & \textbf{14.18} & 0.5236 & 16.81 & 0.4948\\
%  \cline{2-6}
%  & Dual-40\% & & & 15.97 & \textbf{0.5811}\\
%  \cline{2-6}
%  & Tridiagonal-60\% & & & 20.74 & \textbf{0.5822}\\
 \hline\hline
\end {tabular}%
}
\end {table}

%
% Figure: Noise Transition Matrix on MNIST
%
\begin{figure}[!h]
    \centering
     \includegraphics[width=1\textwidth]{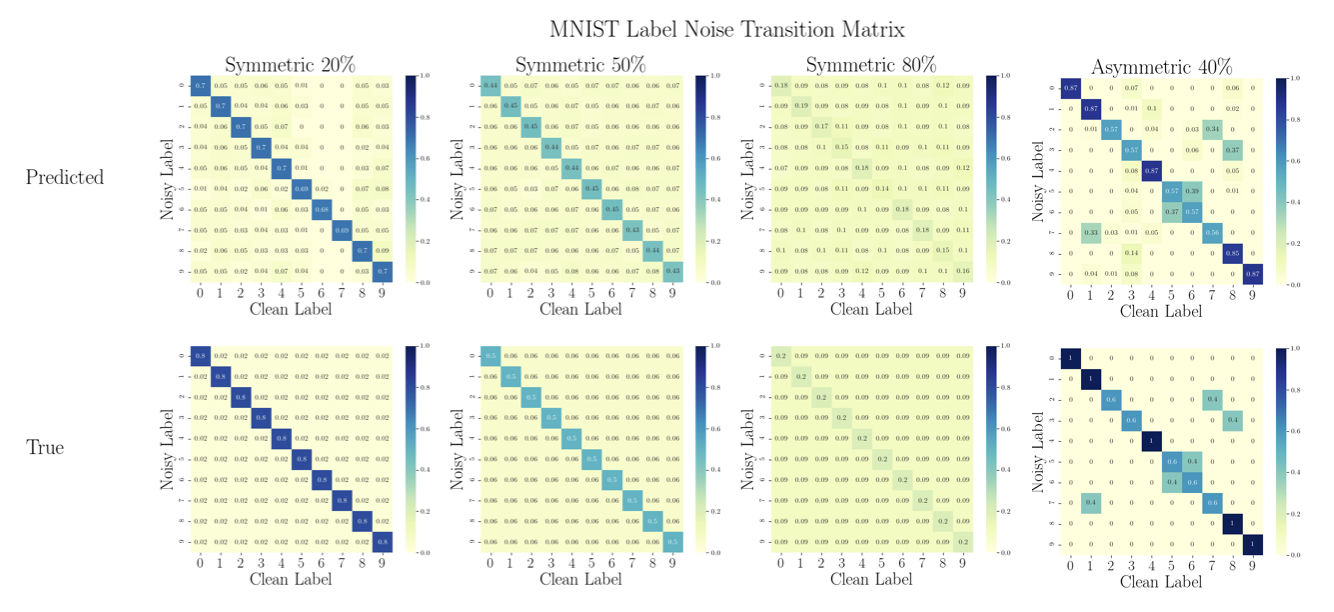}
    \caption{Noise Transition Matrix on MNIST}
    \label{fig:tm_mnist}
\end{figure}

%
% Figure: Noise Transition Matrix on CIFAR10
%
\begin{figure}[!h]
    \centering
    \includegraphics[width=1\textwidth]{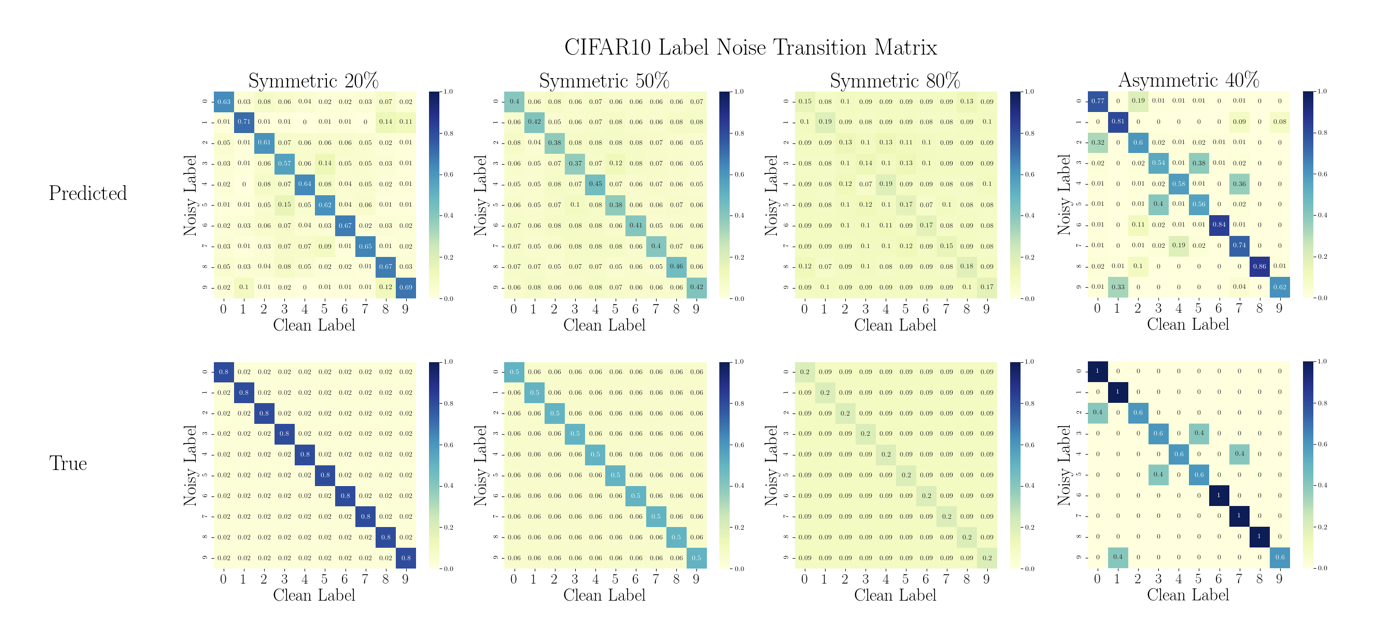}
    \caption{Noise Transition Matrix on CIFAR10}
    \label{fig:tm_cifar10}
\end{figure}

We report the average total variation (ATV) and Kendall Tau rank distance (KTD) \cite{kendall_38} to evaluate the transition matrix estimation, which are defined as follows:

\begin{equation}
    \text{Average total variance} = \frac{1}{C} \sum_{i=1}^C \frac{1}{2} \sum_{j=1}^C {|T_{ij}-\hat{T}_{ij}|}
\end{equation}
\begin{equation}
    \text{Average Kendall Tau Distance} = \frac{1}{C} \sum_{i=1}^C 
    {\sum_{j,k}\bar{K}_{j,k}(\mathbf{t}_i,\hat{\mathbf{t}}_i})
\end{equation}
where $\mathbf{t}_i$ denotes for $i$-th row vector of transition matrix $T$ and $\hat{\mathbf{t}}_i$ for $i$-th row vector of $\widehat{T}$. 

The function $\bar{K}$ for two arbitrary vectors $\mathbf{t}_1$ and $\mathbf{t}_2$ is defined as follows:
\begin{equation}
    \bar{K}_{j,k}(\mathbf{t}_1,\mathbf{t}_2)=
    \begin{cases}
        0 & \text{if j and k are in same order of ranking} \\
        1 & \text{if not}
    \end{cases}
\end{equation}
The total variance is an average l1 norm between two matrices, which denotes the absolute difference between the estimated and ground-truth matrices. Kendall-tau rank distance is defined as a metric that counts the number of pairwise disagreements between two ranking lists, which means comparing the ranking of two matrices.

We measure ATV and KTD with comparison to Noise Adaptation \cite{goldberger_16}. We set compared method to Noise Adaptation since it estimates the noise transition matrix per image, which can be further expanded to Set-Dependent Noise (SDN) settings on Section \ref{sec:sdn_tm}. The result is shown in Table \ref{tab:tm}, indicating that the proposed method outperforms the compared method except for CIFAR10 Asymmetry-40\% with ATV. However, the proposed method outperforms on KTD measure, meaning that the proposed method captures the rank of the ground-truth noise transition matrix more precisely.

\subsubsection{Comparison with Semi-Supervised Method} \label{sec:ssl}

%
% Table
%
\begin {table}[!t]
\caption {Comparison with SSL} \label{tab:dividemix} 
\resizebox{\columnwidth}{!}{%
\begin{tabular}{c c|c|c|c|c}
 \hline \hline
    & Method &Symmetry-20\%& Symmetry-50\% & Symmetry-80\% & Asymmetry-40\%\\
\hline
% \multirow{6}{*}{DirtyMNIST} &
\multirow{4}{*}{CIFAR10}
& DivideMix \cite{li_20} & 85.06 ± 0.09 & 85.08 ± 0.09 & 53.24 ± 0.41 & 76.32 ± 0.17 \\
\cline{2-6}
& MLN (Ours) & 84.20 ± 05 & 77.88 ± 0.07 & 41.83 ± 0.10 & 76.72  ±0.11 \\
\cline{2-6}
& MLN (Ours) + MixUp \cite{zhang_17}& 87.77±0.08 & 84.59±0.09 & 42.38±0.21 & 75.35±0.12 \\
\cline{2-6}
& MLN + DivideMix \cite{li_20} & \textbf{86.82} ± 0.16 &\textbf{88.76} ± 0.11 & \textbf{64.91} ± 0.3 & \textbf{82.40} ± 0.21 \\
\hline
\multirow{4}{*}{CIFAR100}
& DivideMix \cite{li_20} & 62.29 ± 0.17 & \textbf{58.17} ± 0.16 & 40.28 ± 0.17 & 46.83 ± 0.21 \\
\cline{2-6}
& MLN (Ours) & 51.60 ± 0.08 & 42.22±0.07 & 19.88±0.14 & 36.36±0.10\\
\cline{2-6}
& MLN (Ours) + MixUp \cite{zhang_17}& 55.15±0.22 & 46.75±0.17 & 26.17±0.21 & 41.38±0.19 \\
\cline{2-6}
& MLN (Ours) + DivideMix \cite{li_20} & \textbf{62.70} ± 0.05 & 57.68 ± 0.13 & \textbf{40.55} ± 0.24 & \textbf{49.22} ± 0.34\\
\hline \hline

\end {tabular}%
}
\end {table}

For combating noisy labels, the state-of-the-art methods deploy semi-supervised methods with additional augmentation methods, e.g., Dividemix \cite{li_20}. On the other hand, the proposed method focuses on robust architecture and loss function to deal with this problem. As mentioned in Xia et al. \cite{xia_22}, to make the comparison fair, we combine our method with semi-supervised learning. Making the architecture and loss function robust to label noise in a semi-supervised framework will prevent the model from learning the wrong target signal when the estimated clean set is still noisy.

Instead of a linear classification head, we added a Mixture of Logit network heads to predict the label. To fit per-sample loss distribution, we set the selection function as follows
\begin{equation}
    \ell (\theta) = \{\ell_i\}_{i=1}^N = \{\sum_{c=1}^C p(\tilde{y}=c|\mathbf{x}) \log(\hat{p}(y=c|\mathbf{x},z=k;\theta))\}
\end{equation}
where $k = \arg\max_k p(z=k|\mathbf{x};\theta)$.

In addition, for making the pseudo-label, we replace $p_{model}(x;\theta)$ as $\hat{p}(y=c|\mathbf{x},z=k;\theta)$ where $k = \arg\max_k p(z=k|\mathbf{x};\theta)$. For updating the model, we replace cross-entropy loss in labeled set with the mixture cross-entropy loss function with epistemic uncertainty regularizer. The labeled loss is as follows
\begin{equation}
    \mathcal{L}_{\mathcal{X}} =-\frac{1}{|\hat{\mathcal{X}}'|}\sum_{(\mathbf{x},y) \in \hat{\mathcal{X}'} } \left[\sum_{k=1}^K \hat{p}(z=k) \sum_{c=1}^C y^c\log(\hat{p}(y=c|\mathbf{x},z=k;\theta)-\lambda \cdot \sigma_e \right])
\end{equation}
where $\sigma_e$ is same as equation \ref{eq:epis} and we set $\lambda =1$.
We did not learn the $\{\sigma_i\}_{i=1}^K$ in this architecture. This is because as the DivideMix \cite{li_20} framework first has a warm-up phase with a noisy set and then divides the set with an estimated clean labeled set, learning $\{\sigma_i\}_{i=1}^K$ will be unstable as the training set becomes relatively clean after warm-up phase during training.

The clean test set accuracies in CIFAR10 and CIFAR100 datasets are shown in Table \ref{tab:dividemix}. First of all, we observe that using additional data augmentation and regularizer like MixUp \cite{zhang_17} would lead to a performance increase. In addition, on the CIFAR10 dataset, we found out that the hybrid method of DivideMix and MLN outperforms naive DivdeMix with the gap of 1.66\%, 3.58\%, 11.67\%, 6.08\% for each noise setting. In the CIFAR100 dataset, the proposed method outperforms DivideMix except for the Symmetry-50\% noise rate. In Table \ref{tab:dividemix_auroc}, we measure the AUROC of the partitioning clean labeled set and noisy unlabeled set during training. We observe that when the noise is heavy, DivideMix fails to divide the clean set. If we use DivideMix with MLN, then the model tends to learn robustly even when the estimated clean sets are still dirty. We found out that this property leads to better performance.

%
% Table
%
\begin {table}[!t]
\caption {AUROC of Dividing Train set on CIFAR10} \label{tab:dividemix_auroc} 
\resizebox{\columnwidth}{!}{%
\begin{tabular}{c|c|c|c|c}
 \hline \hline
    Method &Symmetry-20\%& Symmetry-50\% & Symmetry-80\% & Asymmetry-40\%\\
\hline
DivideMix \cite{li_20} & 0.979 & \textbf{0.982} & 0.635 & 0.872 \\
\hline
MLN + DivideMix & \textbf{0.987} & 0.977 & \textbf{0.909} & \textbf{0.934} \\
\hline \hline

\end {tabular}%
}
\end {table}

%
% Set-Dependent Noise (SDN) Setting
%
\subsection{Set-Dependent Noise (SDN) Setting}

%
% Figure
%
\begin{figure}[!h]
    \centering
    \includegraphics[width=0.8\textwidth]{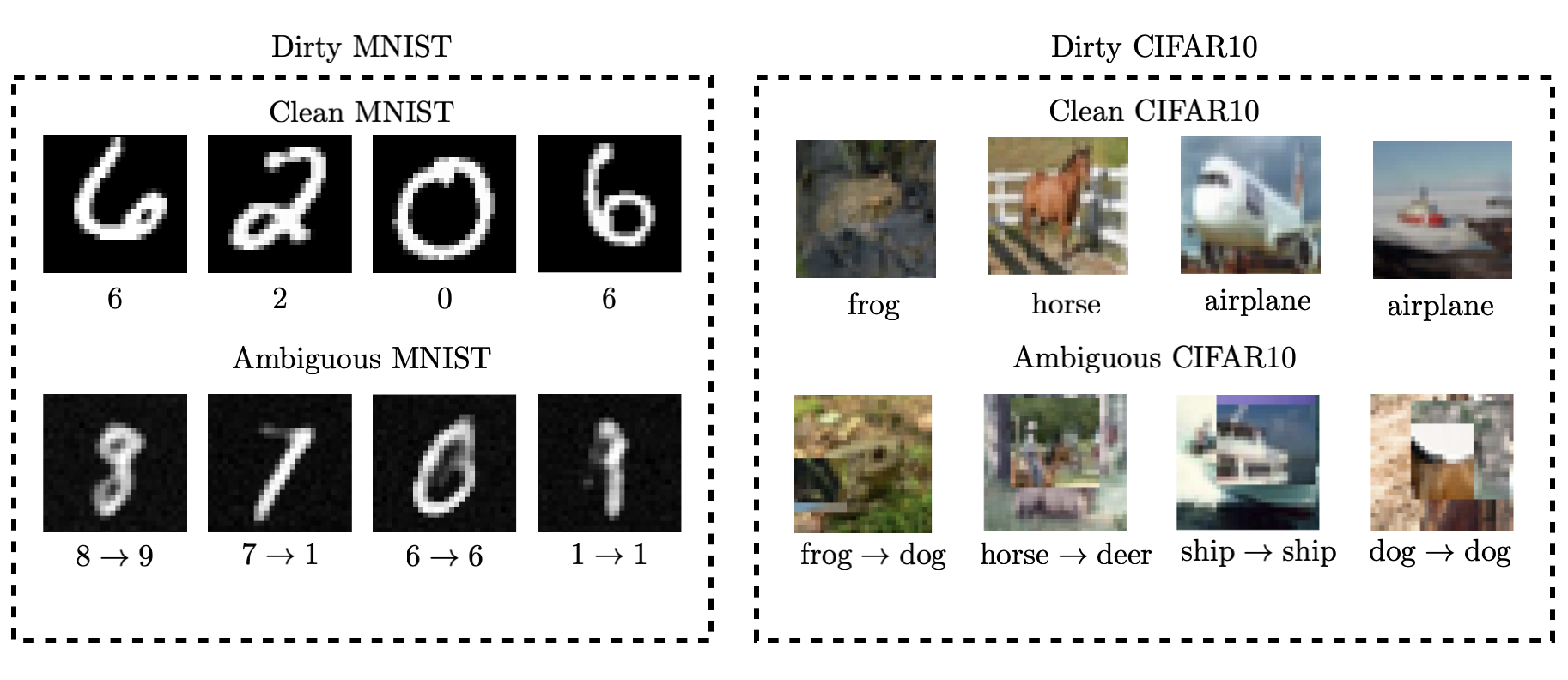}
    \caption{Example of Set-Dependent Noise Dataset. The arrows denote the label corruption.}
    \label{fig:SDN}
\end{figure}

In this section, we show the effectiveness of the MLN on SDN settings. We use the Dirty-MNIST and Dirty-CIFAR10 datasets explained in Figure \ref{fig:SDN}. These datasets contain a clean set and an ambiguous set, where an ambiguous set is composed of ambiguous instances with corrupted labels.
We first show that the robust learning performance on the SDN setting outperforms the previous method, including a semi-supervised method with an additional regularizer like DivideMix \cite{li_20}. Next, we validate proposed uncertainty measure can partition the clean and ambiguous set. Third, we evaluate the estimated noise transition matrix as well. Finally, we experiment on the Instance-Dependent Noise setting, which can be seen as an extreme case of the Set-Dependent Noise setting where the number of the set is the same as the number of instances.

%
% Accuracy
%
\subsubsection{Robust Learning Accuracy}

%
% Table: 
%
\begin {table}[!t]
\caption {Dirty-MNIST Test Accuracy in SDN setting} \label{tab:dirty-mnist_acc} 
\resizebox{\columnwidth}{!}{%
\begin{tabular}{c||c|c|c|c}
\hline
Method & Symmetry-20\% & Symmetry-50\% & Symmetry-80\% & Asymmetry-40\% \\
\hline \hline
Noise Adaptation \cite{goldberger_16}& 79.17 $\pm$ 0.02 & 89.31 $\pm$ 0.03 & 89.14 $\pm$ 0.02 & 79.12 $\pm$ 0.02 \\
\hline
F-correction\cite{patrini_17}& 99.17 $\pm$ 0.04 & 99.14 $\pm$ 0.04 & 99.18 $\pm$ 0.04 & 99.26 $\pm$ 0.02\\
\hline
Co-teaching\cite{han_18} & \textbf{99.32} $\pm$ 0.02 & 99.04 $\pm$ 0.03 & 90.93 $\pm$ 0.57&98.68 $\pm$ 0.07 \\
\hline
Co-teaching+\cite{yu_19} & 98.67 $\pm$ 0.06 & 99.00 $\pm$ 0.06 & 99.06 $\pm$ 0.07 & 98.78 $\pm$ 0.08\\
\hline
JoCoR\cite{wei_20} & 98.81 $\pm$ 0.06 & 98.13 $\pm$ 0.01 & 98.09 $\pm$ 0.0 & 98.77 $\pm$ 0.03\\ 
% \hline
% DivideMix-SGD\cite{li_20}& & & &\\
\hline
DivdeMix\cite{li_20} & 98.11 $\pm$ 0.01 & 98.46 $\pm$ 0.01 & 98.81 $\pm$ 0.0 & 97.94 $\pm$ 0.01 \\ 
\hline
MLN (ours) & 99.26 $\pm$ 0.01 & \textbf{99.45} $\pm$ 0.01 & \textbf{99.40} $\pm$ 0.02 & \textbf{99.31} $\pm$ 0.01\\
\hline \hline
\end {tabular}%
}
\end {table}

%
% Table: 
%
\begin {table}[!t]
\caption {Dirty-CIFAR10 Test Accuracy in SDN setting} \label{tab:dirty_cifar10_acc} 
\resizebox{\columnwidth}{!}{%
\begin{tabular}{c||c|c|c|c}
\hline
Method & Symmetry-20\% & Symmetry-50\% & Symmetry-80\% & Asymmetry-40\% \\
\hline \hline
Noise Adaptation \cite{goldberger_16} & 72.35 $\pm$ 0.06 & 62.94 $\pm$ 0.13 & 72.29 $\pm$ 0.13 & 72.5 $\pm$ 0.09\\
\hline
F-correction\cite{patrini_17} & 81.55 $\pm$ 0.27 & 77.93 $\pm$ 0.4 & 79.74 $\pm$ 0.24 &  82.73 $\pm$ 0.15\\
\hline
Co-teaching\cite{han_18} & 87.64 $\pm$ 0.12 & 83.09 $\pm$ 0.13 & 56.08 $\pm$ 0.14 & 81.33 $\pm$ 0.11\\
\hline
Co-teaching+\cite{yu_19} &86.02 $\pm$ 0.11 & 84.89 $\pm$ 0.16 & 71.55 $\pm$ 0.06 & 85.96 $\pm$ 0.19\\
\hline
JoCoR\cite{wei_20} & \textbf{87.75} $\pm$ 0.08 & 82.72 $\pm$ 0.08 & 48.19 $\pm$ 0.07 &87.46 $\pm$ 0.05\\ 
\hline
DivideMix\cite{li_20}& 80.10 $\pm$ 0.11 & 84.46 $\pm$ 0.15 & 85.34 $\pm$ 0.15 & 73.72 $\pm$ 0.1\\
% \hline
% DivdeMix-Adam\cite{li_20} & 73.8 $\pm$ 0.33 & 80.03 $\pm$ 0.21 & 81.40 $\pm$ 0.18& 68.8 $\pm$ 0.17\\ 
\hline
MLN (ours) & 86.24 $\pm$ 0.06 & \textbf{86.59} $\pm$ 0.07 & \textbf{87.45} $\pm$ 0.05 & \textbf{87.79} $\pm$ 0.04\\
\hline \hline
\end {tabular}%
}
\end {table}

First, we evaluate the accuracy of clean test set of Dirty-MNIST dataset and Dirty-CIFAR10 dataset, shown in Table \ref{tab:dirty-mnist_acc} and  \ref{tab:dirty_cifar10_acc} respectively. We observe that the proposed method outperforms compared methods except for Symmetry-20\% noise rate, where the noise ratio is small. We would like to emphasize that the proposed method works better than the semi-supervised method like DivideMix \cite{li_20} even without any data augmentation like MixUp \cite{zhang_17} or MixMatch \cite{berthelot_19}. This is because, in the SDN setting, the model has to be robust not only to label corruption but also to noisy input as well. MLN obtains its robustness on both label corruption and input noise by mixture-of-experts architecture and its uncertainty. We would like to emphasize that MLN has its strength in robust learning on the corruption of output is dependent on \textit{quality} of input. SDN setting can be more practical in that it is more natural to assume that the annotators will be more likely to make mistakes on a specific subset consisting of hard instances.

%
% Partitioning Sets on the SDN Setting
%
\subsubsection{Partitioning Sets}\label{sec:part_sdn}

%
% Table
%
\begin {table}[!h]
\centering
\caption {Measured AUROC for Partioning Sets} \label{tab:auroc} 
\resizebox{\columnwidth}{!}{%
\begin{tabular}{c c|c|c|c}
 \hline \hline
 & Noise Rate & Noise Adaptation \cite{goldberger_16}& DivideMix \cite{li_20}& MLN (Ours)\\
 \hline
 \multirow{4}{*}{Dirty MNIST} 
 & Symmetry-20\% & 0.3549 & 0.8859 & \textbf{0.9895} \\
 \cline{2-5}
 & Symmetry-50\% & 0.5778 & 0.9161 & \textbf{1.0000}\\ 
  \cline{2-5}
 & Symmetry-80\% & 0.5544 & 0.9269 & \textbf{1.0000}  \\
 \cline{2-5}
 & Asymmetry-40\% & 0.3901 & 0.8714 & \textbf{0.9895} \\
 \hline
 \multirow{4}{*}{Dirty CIFAR10} 
& Symmetry-20\% & 0.5396 & 0.6017 & \textbf{0.9061} \\
 \cline{2-5}
 & Symmetry-50\% & 0.8353 & 0.6378 & \textbf{0.9924}\\ 
 \cline{2-5}
 & Symmetry-80\% & 0.9932 & 0.6751 & \textbf{0.9985}\\ 
 \cline{2-5}
 & Asymmetry-40\% & 0.4500 & 0.5085 & \textbf{0.7117}\\
 \hline\hline
\end {tabular}%
}
\end {table}

Next, we show the ability of the proposed method to partition the dataset leveraging aleatoric uncertainty estimated from the MLN. In particular, the collective outliers, partitions of training data with corrupted labels, are well captured via aleatoric uncertainty. 

We measure AUROC of partitioning two sets, compared with Noise Adaptation \cite{goldberger_16} and state-of-the-art method DivideMix \cite{li_20}. Metric for partitioning set on Noise Adaptation \cite{goldberger_16} is KL divergence between identity matrix and instance wise estimated noise transition matrix, since the estimated noise transition matrix of clean instances will be close to identity matrix if trained properly. In addition, we use soft-max entropy as a metric on DivideMix \cite{li_20}. The result are shown in Table \ref{tab:auroc}. We have found out that the proposed aleatoric uncertainty measure outperforms all of the compared methods, indicating that aleatoric uncertainty from MLN is valid for capturing collective outliers in the noisy datasets.

\begin{figure}[!t]
    \centering
    \includegraphics[width=1\textwidth]{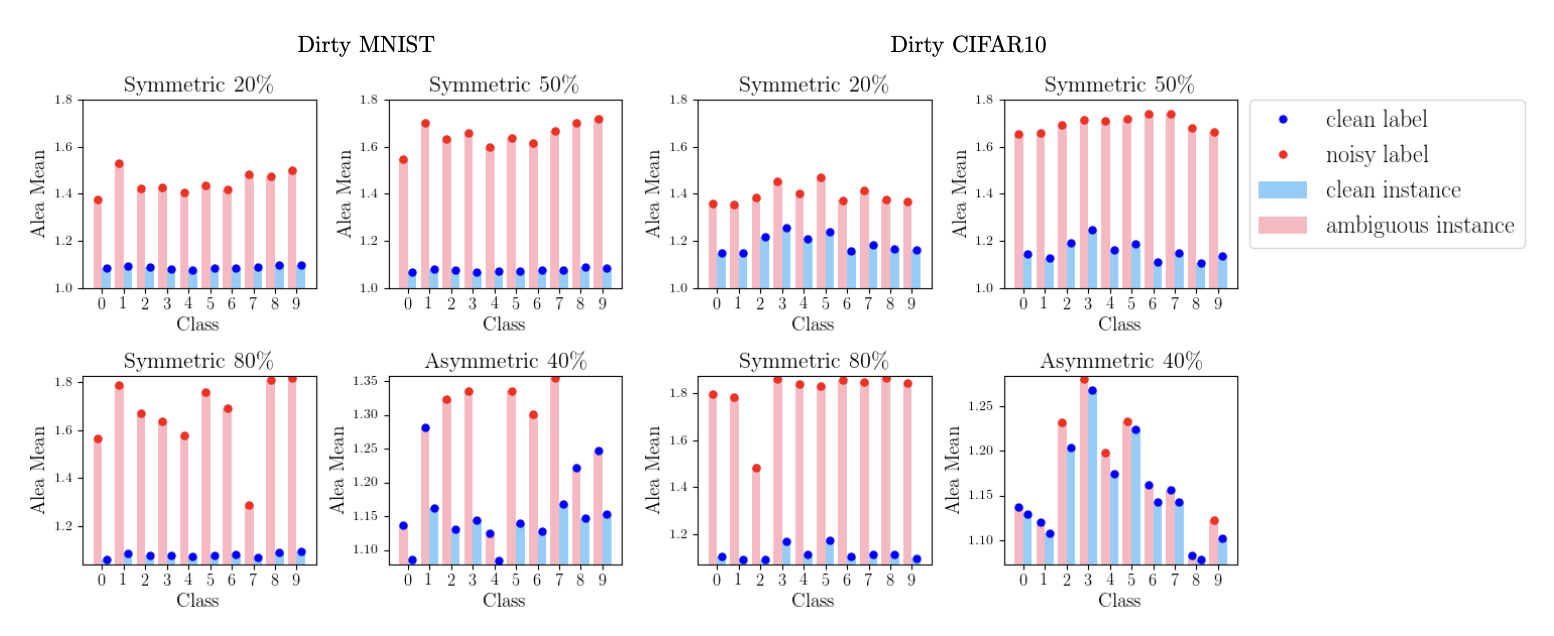}
    \caption{Aleatoric uncertainty along each label on Dirty MNIST and Dirty CIFAR10}
    \label{fig:alea}
\end{figure}

Figure \ref{fig:alea} shows the average of aleatoric uncertainty for both clean and ambiguous sets among each class. The result first shows that the aleatoric uncertainty is higher in ambiguous instances. Furthermore, shown in the symmetric noise cases, a heavy label corruption rate leads to higher aleatoric uncertainty. Finally, Dirty-MNIST with Asymmetry-40\% label noise case shows that aleatoric uncertainty increases in the corrupted labels compared to clean labels in ambiguous instances. This demonstrates that both corruption in instances and corruption in the label are related to aleatoric uncertainty.

%
% Figure
%

%
% Noise Transition Matrix Estimation on the SDN
%
\subsubsection{Noise Transition Matrix Estimation} \label{sec:sdn_tm}

%
% Table
%
\begin {table}[!t]
\caption {Evaluation of Noise Transition Matrix on SDN setting only with estimated ambiguous set \\ Average total variation(x100)(Kendall-Tau distance)} \label{tab:dirty_tm} 
\resizebox{\columnwidth}{!}{%
\begin{tabular}{c c|c|c|c|c}
 \hline \hline
    & Method &Symmetry-20\%& Symmetry-50\% & Symmetry-80\% & Asymmetry-40\%\\
\hline
% \multirow{6}{*}{DirtyMNIST} &
\multirow{6}{*}{Dirty MNIST}
& \multirow{2}{*}{Noise Adaptation\cite{goldberger_16}}  & 25.68 & 45.83 & 73.11& 25.42 \\
& & (0.4472) & (0.4472) & (0.4472) & (0.5164) \\
\cline{2-6}
& \multirow{2}{*}{DivideMix \cite{li_20}} & \textbf{14.00} & 21.12 & 46.88 &  33.02 \\ 
& & (0.4472) & (0.4472) & (0.4472) & (0.4875) \\
\cline{2-6}
& \multirow{2}{*}{MLN (Ours)} & 22.43 & \textbf{13.70}& \textbf{7.99} & \textbf{21.63}\\
& & (0.4472) & (0.4472) & (0.4472) & (0.5326) \\
\hline\hline
\multirow{6}{*}{Dirty CIFAR10}
& \multirow{2}{*}{Noise Adaptation\cite{goldberger_16}} & 29.22 & 49.29 & 66.26 & 20.15 \\
& & (0.4472) & (0.4472) & (0.4472) & (0.5230) \\
\cline{2-6}
& \multirow{2}{*}{DivideMix \cite{li_20}} & \textbf{17.74} & 19.06 & 46.12 &33.60\\ 
& & (0.4472) & (0.4472) & (0.4472)  & (0.5237)\\
\cline{2-6}
& \multirow{2}{*}{MLN (Ours)} & 21.14 & \textbf{14.59} & \textbf{7.94} & \textbf{18.93}\\
& &  (0.4472) & (0.4472) & (0.4472) & (0.5090) \\
\hline \hline

\end {tabular}%
}
\end {table}

%
% Figure: Noise Transition Matrix on Dirty-MNIST
%
\begin{figure}[!h]
    \centering
    \includegraphics[width=1\textwidth]{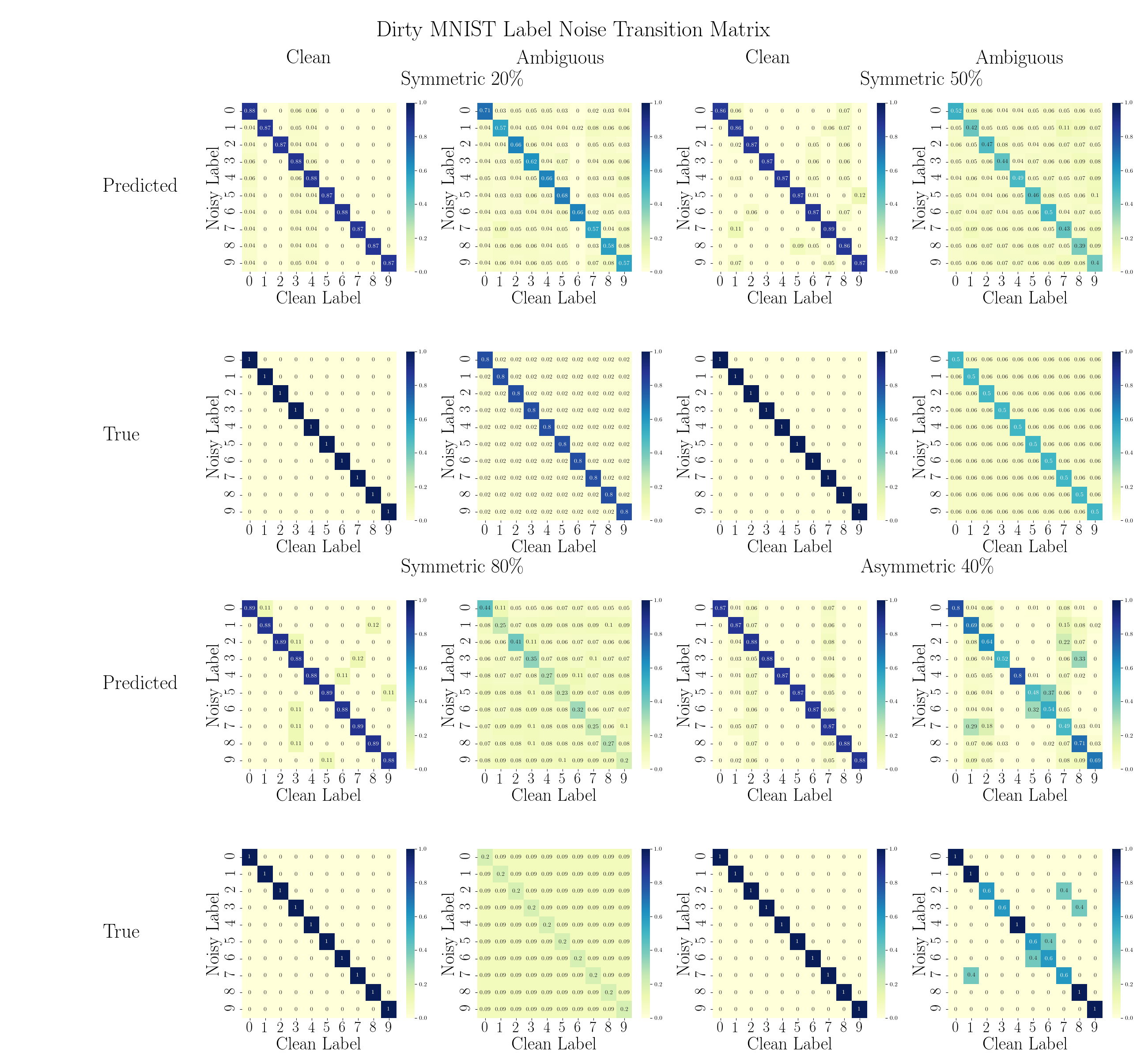}
    \caption{Noise Transition Matrix on Dirty-MNIST}
    \label{fig:tm_dirtymnist}
\end{figure}

%
% Noise Transition Matrix on Dirty-CIFAR10
%
\begin{figure}[!h]
    \centering
    \includegraphics[width=1\textwidth]{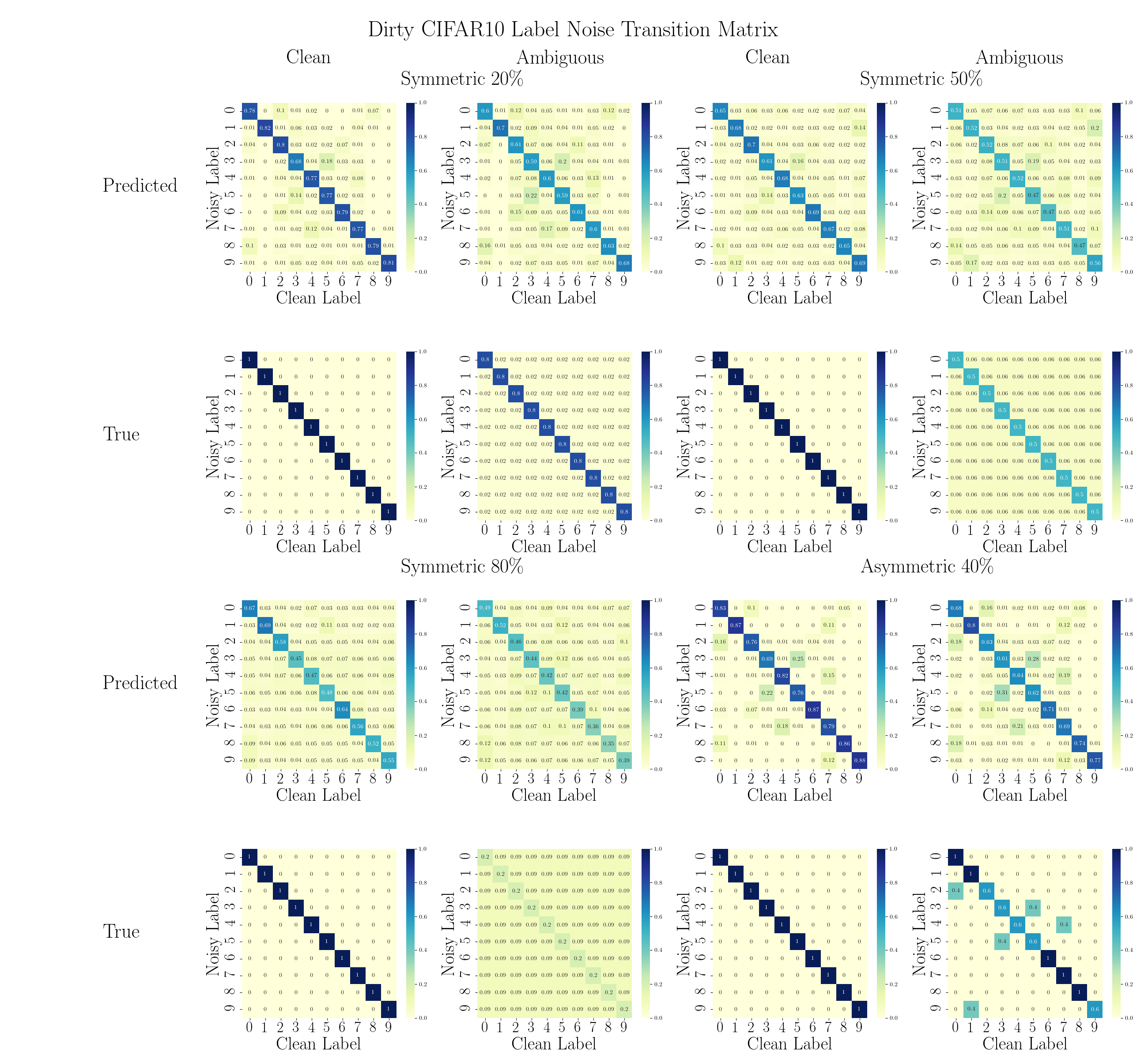}
    \caption{Noise Transition Matrix on Dirty-CIFAR10}
    \label{fig:tm_dirtycifar10}
\end{figure}

We then examine the ability of our method to estimate noise transition matrices per each group where we partition the training data into multiple sets using the predicted aleatoric uncertainty in Section \ref{sec:part_sdn}.
We first evaluate Average total variance and Kendall-Tau distance on \textit{estimated ambiguous set} comparing with Noise-Adaptation \cite{goldberger_16} and DivideMix \cite{li_20} same as Section \ref{sec:part_sdn}. For Noise-Adaptation \cite{goldberger_16}, we average the estimated noise transition matrix on the estimated set. For DivideMix \cite{li_20} we average the softmax output to obtain the estimated noise transition matrix since the softmax output can be seen as a confidence score that can be interpreted as a noise ratio. We cannot use a confusion matrix for this setting because the noise transition matrix is estimated on an unseen test set. The noise transition matrix is defined as follows:

\begin{equation}
    T_{i,j}^{\text{DivideMix}} = \hat{P}(y=j|\mathbf{x},f(\mathbf{x})=i) \label{eq:dividemix_tm}
\end{equation}
where $f(\mathbf{x})$ is a prediction, $\hat{P}$ is a soft-max output.

The experimental results shown in Table \ref{tab:dirty_tm}. Except for Symmetry-20\% on the Dirty-MNIST dataset, the proposed method outperforms the compared method. The gap increases as the noise ratio increases since other methods have a tendency to have overconfident estimates even if the noise ratio is high. 

Furthermore, the estimated noise transition matrix is illustrated in Figure \ref{fig:tm_dirtymnist} and \ref{fig:tm_dirtycifar10} on the Dirty-MNIST and Dirty-CIFAR10 datasets respectively.
Each quarter of the figure denotes a single experiment for each corruption pattern, with the upper and lower rows showing the predicted noise transition matrix and the ground truth, respectively. Here, clean labels on the ambiguous set denote the ground-truth label of each instance after ambiguating the instances. We can see that our proposed method is able to correctly estimate the noise transition matrix for both clan and ambiguous sets in terms of a row-wise ranking manner. Furthermore, Figure \ref{fig:tm_dirtycifar10} suggests that our proposed method can also capture the noise-induced in inputs (i.e., CutMix~\cite{yun_19}). In other words, the images of cats are cut-mixed with the images of dogs (and vice versa), and these corruption patterns are well captured by the noise transition matrix. 

%
% IDN
%
\subsubsection{Instance-Dependent Noise}

We further experiment on the Instance-Dependent Noise setting, which can be seen as an extreme version of the Set-Dependent Noise (SDN) setting where the number of the set is the same as the number of instances. We measure clean test accuracy on noise ratio of 20\% and 40\% on MNIST and CIFAR10 datasets, respectively. The result are shown in Table \ref{tab:idn_acc}. 
%
% Table: 
%
\begin {table}[!t]
\centering
\caption {Instance-Dependent Noise (IDN) Setting} \label{tab:idn_acc} 
\resizebox{\columnwidth}{!}{%
\begin{tabular}{|c||c|c|c|c|}
\hline
& \multicolumn{2}{c}{MNIST} \vline & \multicolumn{2}{c}{CIFAR10} \vline \\
 \cline{2-5}
Noise Rate & 20\% & 40\% &  20\% & 40\% \\
\hline
Noise Adaptation \cite{goldberger_16} & \textbf{99.24} $\pm$ 0.03 & 91.03 $\pm$ 0.86 & 68.9 $\pm$ 0.2 & 45.8 $\pm$ 0.14\\
\hline
F-correction\cite{patrini_17} & 89.65 $\pm$ 1.54 & 68.66 $\pm$ 0.68 & 71.84 $\pm$ 2.74 & 48.58 $\pm$ 1.43\\
\hline
Co-teaching\cite{han_18} & 98.03 $\pm$ 0.31 & 95.31 $\pm$ 0.49 & 82.5 $\pm$ 0.11 & 61.33 $\pm$ 0.28\\
\hline
Co-teaching+\cite{yu_19} & 98.57 $\pm$ 0.14 & 98.58 $\pm$ 0.07 & \textbf{85.79} $\pm$ 0.16 & 33.76 $\pm$ 0.11 \\
\hline
JoCoR\cite{wei_20} & 98.94 $\pm$ 0.11 & 98.46 $\pm$ 0.25 & 85.45 $\pm$ 0.14 & 55.88 $\pm$ 0.23 \\ 
\hline
DivdeMix\cite{li_20}& 99.05 $\pm$ 0.03 &\textbf{98.97} $\pm$ 0.03 & 83.77 $\pm$ 0.27 & 61.03 $\pm$ 2.01 \\ 
\hline
MLN (ours) & 98.36 $\pm$ 0.0 & 92.65 $\pm$ 0.01 & 77.18 $\pm$ 0.23&  55.38 $\pm$ 0.11 \\
\hline
MLN + MixUp (ours) & 98.58 $\pm$ 0.03 & 96.44 $\pm$ 0.12 & 85.09 $\pm$ 0.58 & \textbf{61.44} $\pm$ 0.16 \\
\hline \hline
\end {tabular}
}
\end {table}

We observe that MLN with MixUp augmentations has a significant performance increase compared to MLN without any additional augmentations. Although the proposed method does not outperform on noise settings except for CIFAR10 40\%, the gap is small. At MNIST 20\%, the gap is 0.66\%, 2.53 \% for 40\% and 0.60\% for CIFAR 20\%.

\subsection{Experiments on Real-World Dataset}
In this section, We evaluate the MLN on a real-world dataset, i.e., Clothing1M. We first show the test accuracy compared with related work. Then, we qualitatively show the proposed uncertainty can divide the whole test set into clean and ambiguous sets. 
%
% Accuracy
%
% \paragraph{Robust Learning Accuracy}
%
% Table: 
%
\begin {table}[!t]
\caption {Clothing 1M Test Accuracy} \label{tab:cloting1m_acc} 
\resizebox{\columnwidth}{!}{%
\begin{tabular}{c|c|c|c|c|c|c}
\hline
Noise Adaptation \cite{goldberger_16} & F-correction\cite{patrini_17} & Co-teaching\cite{han_18} & 
Co-teaching+\cite{yu_19} & JoCoR\cite{wei_20}  & MLN (ours) \\
\hline 
 67.3 $\pm$ 0.12 & 68.0 $\pm$ 0.10 & 70.33 $\pm$ 0.12 & 68.85 $\pm$ 0.35 & \textbf{71.92} $\pm$ 0.14 &  71.56 $\pm$ 0.05 \\
\hline \hline
\end {tabular}%
}
\end {table}
 
\begin{figure}[!t]
\centering
%\begin{subfigure}{\textwidth}
%    \includegraphics[width=.45\textwidth]{low_alea.png}
%\end{subfigure}
%\hfill
%\begin{subfigure}{\textwidth}
%    \includegraphics[width=.45\textwidth]{top_alea.png}
%\end{subfigure}

\subfigure{
\includegraphics[width=.45\columnwidth]{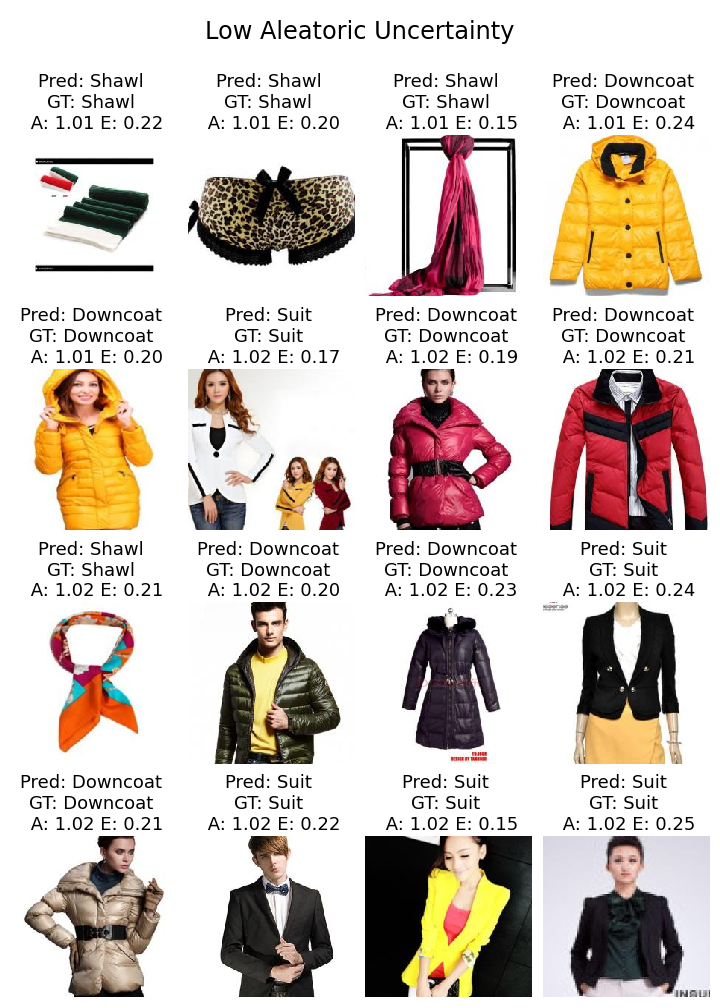}
}
\hfill
\subfigure{
\includegraphics[width=.45\columnwidth]{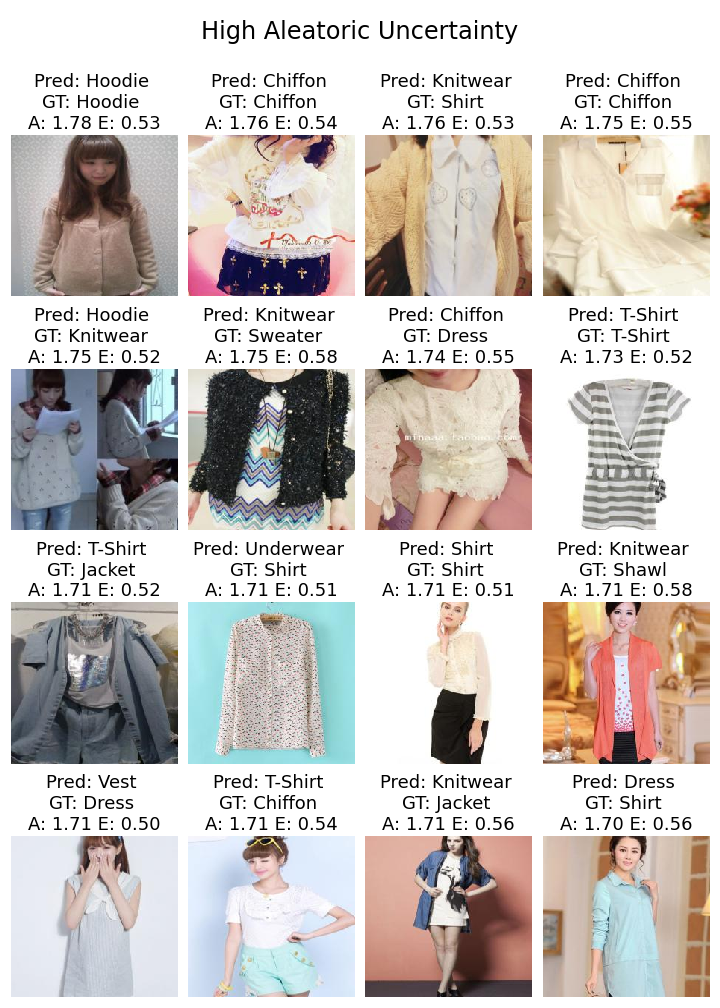}
}
\caption{
Clothing1M test set samples with low aleatoric uncertainty (left) and high aleatoric uncertainty (right). Pred denotes the predicted label and GT for the ground-truth label. 'A' denotes aleatoric uncertainty, and 'E' denotes epistemic uncertainty.
}
\label{fig:clothing1m_alea}
\end{figure}

\begin{figure}[!t]
\centering
%\begin{subfigure}{\textwidth}
 %   \includegraphics[width=.45\textwidth]{low_epis.png}
%\end{subfigure}
%\hfill
%\begin{subfigure}{\textwidth}
 %   \includegraphics[width=.45\textwidth]{top_epis.png}
%\end{subfigure}
\subfigure{
\includegraphics[width=.45\columnwidth]{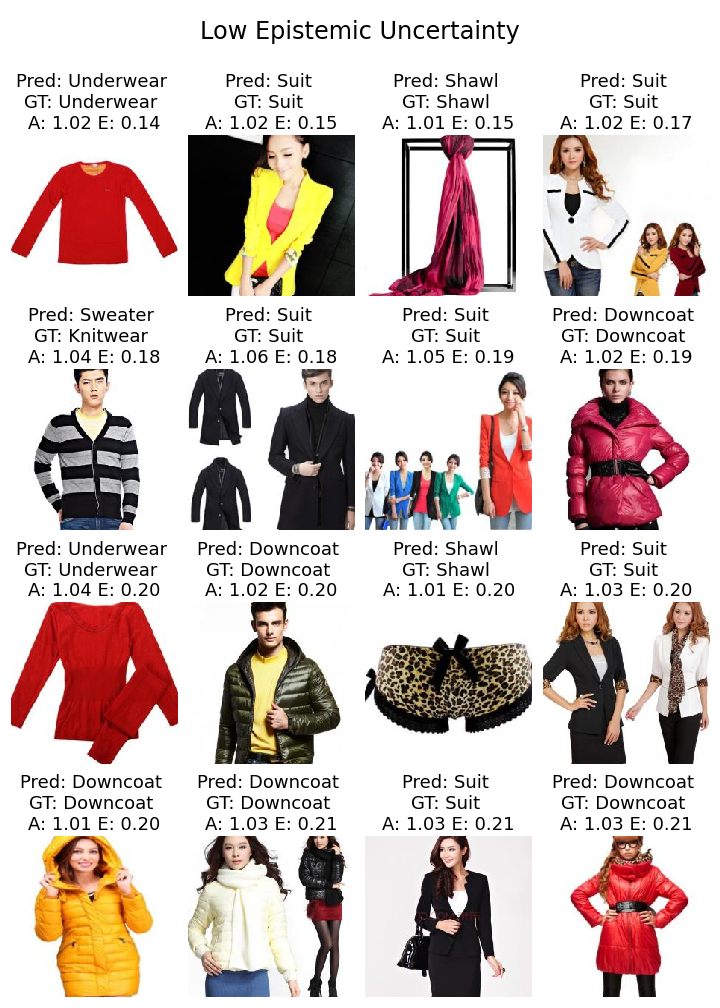}
}
\hfill
\subfigure{
\includegraphics[width=.45\columnwidth]{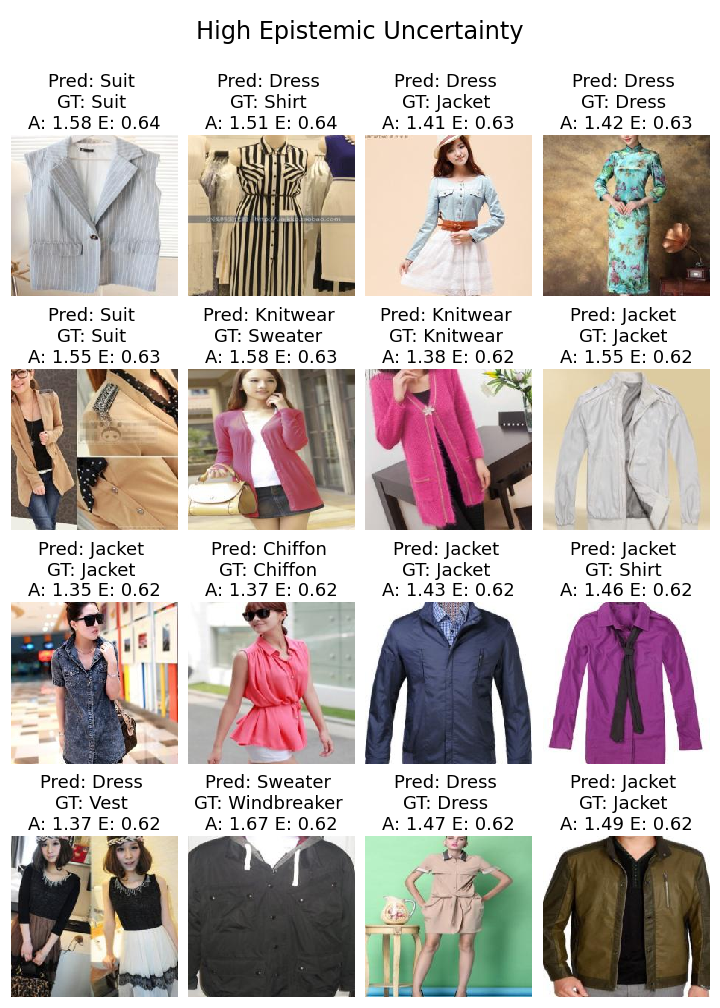}
}
\caption{Clothing1M test set samples with low epistemic uncertainty (left) and high epistemic uncertainty (right). Pred denotes the predicted label and GT for the ground-truth label. 'A' denotes aleatoric uncertainty, and 'E' denotes epistemic uncertainty.
}
\label{fig:clothing1m_epis}
\end{figure}

We measure the test set accuracy, shown in Table \ref{tab:cloting1m_acc}. The proposed method works second-best compared to other methods with a 0.36\% gap. However, some instances on test set shown in Figure \ref{fig:clothing1m_alea} and \ref{fig:clothing1m_epis} are still ambiguous and noisy. From this, we would like to claim that this minor gap is not significant in measuring the performance of robust learning since there still exists some noisy instances and labels in the test set.

%
% Uncertainty
%
% \paragraph{Qualitative Results on Uncertainty}

%
% Figure
%

%
% Figure
%

% alea는 single object, centered 를 잘 잡아준다. epis는 오히려 모호하게 보이는것? 여러가지로 해석가능한거

% alea가 높은거는 옷을 여러개를 같이 입음.  1.3 보면 ... (레이어드) 
% epis는 애초에 두가지로 해석가능한거. 

% 공통적으로 나타나는 거
We then sample instances from the test set, with high uncertainty instances and low uncertainty. We sample top 15 and low 15 instances for aleatoric uncertainty and epistemic uncertainty respectively, shown in Figure \ref{fig:clothing1m_alea} and \ref{fig:clothing1m_epis}. We first observe that set with low aleatoric uncertainty and with low epistemic uncertainty has a clean background with a relatively clean image in common. However, both a set with high aleatoric and a set with epistemic uncertainty contains some wrongly predicted instances and relatively complicated images.
% 두개의 다른점 강조
We observe that low aleatoric uncertainty can capture the images containing a single object and are well centered. In addition, images with multiple clothes, i.e., wearing T-shirts inside the jacket on the fourth row in the first column, tend to have high aleatoric uncertainty. On the other hand, instances that can be interpreted as two categories, i.e., shirted style dress on the first row in the second column, tends to report high epistemic uncertainty.

\subsection{Ablation Studies}
In this section, we conduct ablation studies on MLN.
We first show how the performance of MLN varies with uncertainty regularizers ($\lambda_1, \lambda_2$) on the Dirty-CIFAR10 dataset. 
Then, we explore how other uncertainty measures perform in partitioning sets on SDN settings.

\subsubsection{Effects of $\lambda$}

%
% Table: 
%
\begin {table}[!t]
\centering
\caption {Ablation Study on $\lambda$ on SDN CIFAR10 } \label{tab:reg} 
\resizebox{\columnwidth}{!}{%
\begin{tabular}{|c|c||c|c|c|c|c|c|}
\hline
& &\multicolumn{3}{c}{Symmetry 50\%} \vline & \multicolumn{3}{c}{Asymmetry 40\%} \vline \\
 \cline{3-8}
 $\lambda_1$ & $\lambda_2$ & ACC & ATV & AUROC & ACC & ATV & AUROC \\
\hline
$\lambda_1=1$ & $\lambda_2=1$ & \underline{86.59} $\pm$ 0.07 & \textbf{14.59} & \underline{0.9924} & \textbf{87.79} $\pm$ 0.04 & \textbf{18.93} & \underline{0.7117}\\
\hline
$\lambda_1=0.1$ & $\lambda_2=1$ & \textbf{87.15} $\pm$ 0.07 & 49.38 & \textbf{0.9964} & \underline{87.34} $\pm$ 0.05 & \underline{20.04} & \textbf{0.7385} \\
\hline
$\lambda_1=10$ & $\lambda_2=1$ & 76.35 $\pm$ 0.11 & NaN & 0.9844 & 77.83 $\pm$ 0.06 & NaN & 0.6665\\
\hline
$\lambda_1=1$ & $\lambda_2=0.1$ & 85.53 $\pm$ 0.05 & \underline{26.79}& 0.5828 & 86.74 $\pm$ 0.04 & 63.20 & 0.6206 \\
\hline
$\lambda_1=1$ & $\lambda_2=10$ & 85.42 $\pm$ 0.03 & 39.41 & 0.9515 & 87.2 $\pm$ 0.07 & 19.75 & 0.5284\\
\hline \hline
\end {tabular}%
}
\end {table}
We show how performance varies on  different uncertainty regularizer ($\lambda_1, \lambda_2$), on Dirty-CIFAR10 dataset with Symmetry-50\% and Asymmetry-40\% noise ratio. The result is shown in Table \ref{tab:reg}, the value NaN happens when there do not exist any instances with the particular row. We found out that decreasing $\lambda_1$ slightly increases the accuracy of the model with the cost of degrading the performance of noise transition matrix estimation. On the other hand, increasing $\lambda_1$ may harm the model to capture the clean target signal. In addition, decreasing $\lambda_2$ decrease the accuracy of the model slightly and fails to estimate the noise distribution. Increasing $\lambda_2$ also drops the model's accuracy slightly and fails to partition a clean and ambiguous set. As a result, we propose best choice is $\lambda_1 = 1$, $\lambda_2 = 1$.

\subsubsection{Partioning set with different uncertainties in SDN}
There exist different uncertainty measures than the ones explained in (\ref{eq:epis}) and (\ref{eq:alea}), for example, it can be measured by max-softmax \cite{hendrycks_16}, softmax entropy, or from the entropy of mixture weights. 
The Pi-entropy denotes the entropy of the mixture weights. As bigger entropy can be interpreted as a lower weight of target distribution, this can be seen as the uncertainty of the inputs.
In this experiment, we assume the size of two sets is the same and set threshold as the median of the uncertainty measures to partition these sets. The uncertainty measures are as follows.
\begin{equation}
    \text{max-softmax} = 1 - \underset{c}{\text{max}} \ \mu_k^{(c)}(x) \quad \text{where} \quad k=\underset{i}{\text{argmax}} \ \pi_i(x)
\end{equation}
\begin{equation}
    \text{softmax-entropy} = - \sum_c^C \mu_k^{(c)}(x)\log\left(\mu_k^{(c)}(x)\right) \quad \text{where} \ k=\underset{i}{\text{argmax}} \ \pi_i(x)
\end{equation}
\begin{equation}
    \text{pi-entropy} = - \sum_k^K \pi_k(x)\log\left(\pi_k(x)\right)
\end{equation}

%
% Table
%
\begin {table}[!h]
\centering
\caption {Measured AUROC over different uncertainty measures} \label{tab:auroc_ab} 
\resizebox{\columnwidth}{!}{%
\begin{tabular}{c c|c|c|c|c|c}
 \hline \hline
 & Noise Rate & Aleatoric & Epistemic & $\pi$ Entropy & Max Softmax & Softmax Entropy \\
 \hline
 \multirow{4}{*}{Dirty MNIST} 
 & Symmetry-20\% & \textbf{0.9998} & 0.9986 & 0.9935 & 0.9930 & 0.9958\\
 \cline{2-7}
 & Symmetry-50\% & \textbf{1.0000} & 0.9999 & 0.9996 & 0.9957 & 0.9994\\ 
  \cline{2-7}
 & Symmetry-80\% & \textbf{1.0000} & 1.0000 & 0.9993 & 0.9959 & 0.9988 \\
 \cline{2-7}
 & Asymmetry-40\% & 0.9895 & 0.9605 & 0.9581 & 0.9898 & \textbf{0.9909} \\
 \hline
 \multirow{4}{*}{Dirty CIFAR10} 
 & Symmetry-20\% & 0.9061 & 0.6718 & 0.7614 & 0.9026 & \textbf{0.9130}\\
 \cline{2-7}
 & Symmetry-50\% & \textbf{0.9924} & 0.8913 & 0.9600 & 0.9588 & 0.9729\\ 
 \cline{2-7}
 & Symmetry-80\% & \textbf{0.9985} & 0.9896 & 0.9906 & 0.9449 & 0.9765\\ 
 \cline{2-7}
 & Asymmetry-40\% & \textbf{0.7117} & 0.5747 & 0.6364 & 0.7100 & 0.7047\\
 \hline\hline
\end {tabular}%
}
\end {table}

As the observation noise is an exemplar case of aleatoric uncertainty, instance corruption patterns can be captured by aleatoric uncertainty. Table \ref{tab:auroc_ab} reports the aleatoric uncertainty measure can partition clean and ambiguous sets on symmetric noise compared to other uncertainty measures except for Symmetry-20\% noise rate in Dirty-CIFAR10 and Asymmetry-40\% noise rate in Dirty-MNIST, but with a small gap. We observe that the measure AUROC increases as the noise rate increases, indicating higher noise on the label can also increase uncertainty.
% The asymmetric noise max-softmax or softmax-entropy method works better since the MLN makes underconfident estimates on asymmetric noise patterns.
% as asymmetric noise patterns can be seen bias of label distribution, max-softmax or softmax-entropy can be more capable as aleatoric uncertainty denotes the variance of data.

%
% Limitation
%
\subsection{Limitations} \label{subsec:lim}

In this section, we will discuss the limitations of the proposed method. 
While combining the proposed method with the semi-supervised method result in an improvement in the classification performance, the proposed method does not perform as effectively as the semi-supervised method on CCN settings. In addition, the gap of clean test set accuracy is small with JoCoR \cite{wei_20} in the CCN setting except for symmetric 80\% noise rate. However, we observe that further combining the small-loss selection method with small modification or additional augmentation like MixUp \cite{zhang_17} can improve the performance. We would like to claim that the proposed method is complementary to the small-loss selection methods to gain further robustness when the selected small-loss set is still noisy.  Furthermore, although the proposed method does not outperforms state-of-the-art method on CCN and IDN settings, we would like to claim that SDN setting are more pratical than CCN or IDN, because the annotator tends to mislabel the corrupted or ambiguous data.
%
% Conclusion
% 
\section{Conclusion}
% CCN
We have introduced an uncertainty-aware robust learning framework by leveraging a mixture of the experts' model named mixture logit networks (MLN). The MLN can estimate two different types of uncertainty, epistemic and aleatoric, where the predictive uncertainty is further utilized to define a novel regularization method. We showed that the MLN could represent multi-modal distributions, making the model not only robust to outliers but also able to estimate noise patterns. In addition, we found out that combining the proposed method with a semi-supervised small loss selection method can lead to further improvement.
 % SDN
In particular, we presented a Set-Dependent Noise (SDN) learning problem where multiple corruption patterns exist per partition and proposed a novel validation scheme for estimating the corruption patterns. To tackle this problem, we leveraged aleatoric uncertainty to detect the corrupted partition and estimated the SDN patterns using the multi-modal target distribution computed from the MLN. We would like to note that uncertainty estimation on the robust learning framework plays a significant role in providing information about the corruption of each instance.
% Future Work
The current evaluation scheme for the SDN setting relies on two assumptions: the collective outliers can be separated via the estimated aleatoric uncertainty, and a particular label transition matrix exists per each partition. One promising future research direction could be examining our proposed method to real-world datasets without applying artificial noises to both inputs and outputs. 

% Clothing1M dataset can be used for this purpose as it is known to contain a nontrivial amount of mislabeled data. 

\section*{Acknowledgement}
This work was supported by Institute of Information \& communications Technology Planning \& Evaluation (IITP) grant funded by the Korea government(MSIT) (No. 2019-0-00079 , Artificial Intelligence Graduate School Program(Korea University))

%
% bibliography
%
\clearpage
\newpage
\bibliographystyle{elsarticle-num}

\end{sloppypar}
\end{document}